\documentclass[runningheads]{llncs}
\usepackage{graphicx}
\usepackage{comment}
\usepackage{amsmath,amssymb} 
\usepackage{color}

\usepackage{epsfig}
\usepackage{graphicx}
\usepackage{amsmath}
\usepackage{amssymb}
\usepackage{mathrsfs}
\usepackage{url}
\usepackage{enumitem}
\usepackage{algorithm}
\usepackage{algorithmic}
\usepackage{dsfont}
\usepackage{multirow}
\usepackage{booktabs}
\usepackage[pagebackref=false,breaklinks=true,letterpaper=true,colorlinks,urlcolor=blue,citecolor=blue,linkcolor=blue,bookmarks=false]{hyperref}

\newcommand{\figdir}{figures}

\begin{document}
\pagestyle{headings}
\mainmatter
\def\ECCVSubNumber{2761}  

\title{Rethinking Image Deraining via \\ Rain Streaks and Vapors} 

\titlerunning{Rethinking Image Deraining via Rain Streaks and Vapors}
%
%
\author{Yinglong Wang$^1$ \and
Yibing Song$^2$\thanks{Y. Song and B. Zeng are the corresponding authors. The results and code are available at \url{https://github.com/yluestc/derain}.} \and Chao Ma$^3$ \and
Bing Zeng$^{1\star}$}

\authorrunning{Wang et al.}
%
\institute{University of Electronic Science and Technology of China \and Tencent AI Lab \and MoE Key Lab of Artificial Intelligence, AI Institute, Shanghai Jiao Tong University \\
\email{ylwanguestc@gmail.com~~~~~~yibingsong.cv@gmail.com \\ chaoma@sjtu.edu.cn~~~~~~eezeng@uestc.edu.cn}}
\maketitle

\begin{abstract}
Single image deraining regards an input image as a fusion of a background image, a transmission map, rain streaks, and atmosphere light. While advanced models are proposed for image restoration (i.e., background image generation), they regard rain streaks with the same properties as background rather than transmission medium. As vapors (i.e., rain streaks accumulation or fog-like rain) are conveyed in the transmission map to model the veiling effect, the fusion of rain streaks and vapors do not naturally reflect the rain image formation. In this work, we reformulate rain streaks as transmission medium together with vapors to model rain imaging. We propose an encoder-decoder CNN named as SNet to learn the transmission map of rain streaks. As rain streaks appear with various shapes and directions, we use ShuffleNet units within SNet to capture their anisotropic representations. As vapors are brought by rain streaks, we propose a VNet containing spatial pyramid pooling (SSP) to predict the transmission map of vapors in multi-scales based on that of rain streaks. Meanwhile, we use an encoder CNN named ANet to estimate atmosphere light. The SNet, VNet, and ANet are jointly trained to predict transmission maps and atmosphere light for rain image restoration. Extensive experiments on the benchmark datasets demonstrate the effectiveness of the proposed visual model to predict rain streaks and vapors. The proposed deraining method performs favorably against state-of-the-art deraining approaches.
\keywords{Deep Image Deraining}
\end{abstract}

\section{Introduction}

Rain image restoration produces visually pleasing background (i.e.,scene content) and benefits recognition systems (e.g., autonomous driving). Attempts~\cite{jiang-cvpr17-novel,fu-cvpr17-removing} of image deraining formulate rain image as the combination of rain streaks and background. These methods limit their restoration performance when the rain is heavy. The limitation occurs because heavy rain consisting of rain streaks and vapors causes severe visual degradation. When the rain streaks are clearly visible, a part of them accumulate to become vapors. The vapors produce the veiling effect which decreases image contrast and causes haze. Fig.~\ref{fig:intro} shows an example. Without considering vapors, existing deraining methods do not perform well to restore heavy rainy images as shown in Fig.~\ref{fig:intro}(b) and Fig.~\ref{fig:intro}(c).

\renewcommand{\tabcolsep}{1pt}
\def\swthree{0.33\linewidth}
\begin{figure*}[t]
\begin{center}
\begin{tabular}{ccc}
\includegraphics[width=\swthree]{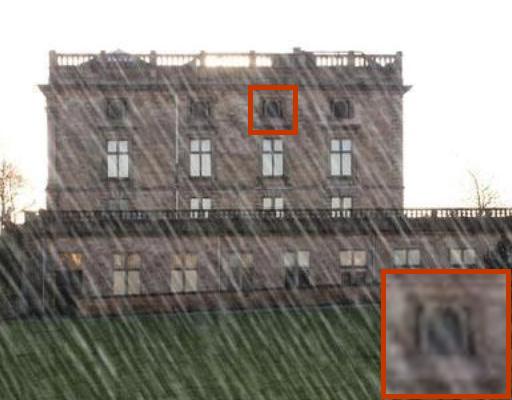}&
\includegraphics[width=\swthree]{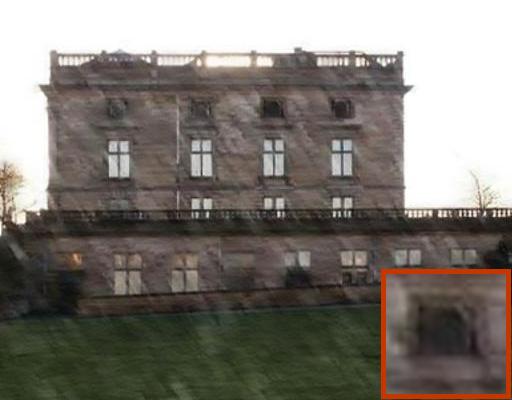} & \includegraphics[width=\swthree]{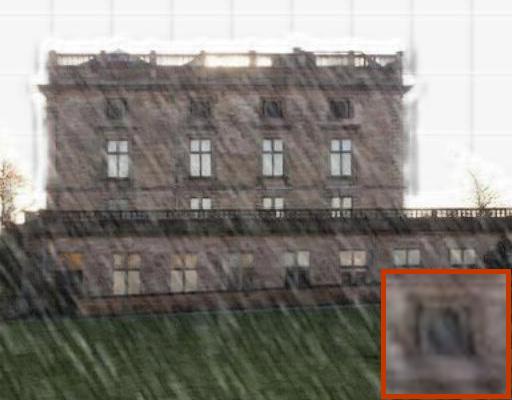} \\
(a) Input & (b) SPANet~\cite{wang-cvpr19-spatial} & (c) JORDER~\cite{yang-pami19-joint} \\
\includegraphics[width=\swthree]{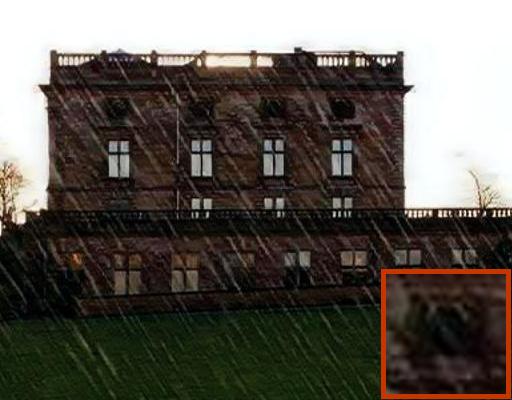} & \includegraphics[width=\swthree]{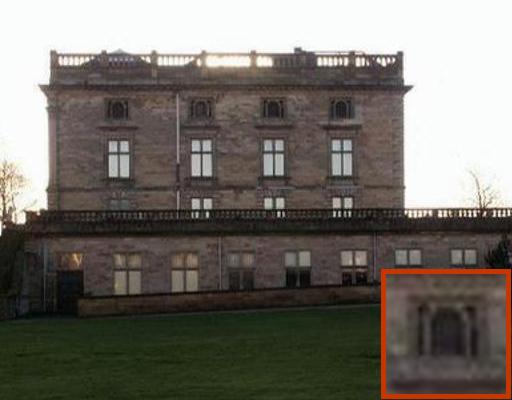}&
\includegraphics[width=\swthree]{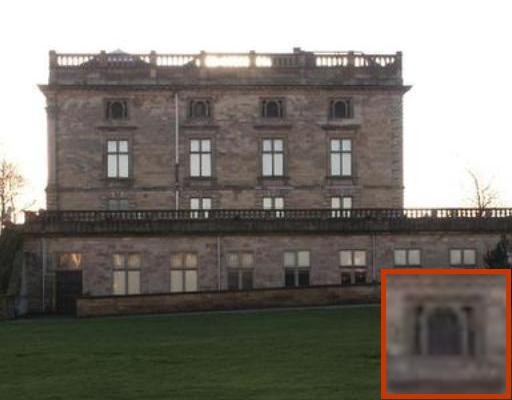} \\
 (d) PYM+GAN~\cite{li-cvpr19-heavy} & (e) Ours & (f) Ground Truth \\
\end{tabular}
\end{center}
\caption{Rain image restoration results. Input image is shown in (a). Results of SPANet~\cite{wang-cvpr19-spatial}, JORDER~\cite{yang-pami19-joint}, PYM+GAN~\cite{li-cvpr19-heavy} are shown in (b)-(d). Ground Truth is shown in (f). The proposed visual model is effective to formulate rain streaks and vapors, which brings high quality deraining result as shown in (e).}
\label{fig:intro}
\end{figure*}

Recent study~\cite{li-cvpr19-heavy} reformulates rain image generation via the following model:
\begin{equation}\label{eq:rain-py1}
  \mathbf{I} = \mathbf{T} \odot (\mathbf{J} + \sum^{n}_{i=1} \mathbf{S}_{i}) + (\mathds{1}-\mathbf{T}) \odot \mathbf{A}
\end{equation}
where $\mathbf{I}$ is the rain image, $\mathbf{T}$ is the transmission map, $\mathbf{J}$ is the background to be recovered, $\mathbf{S}_{i}$ is the rain streak layer, and $\mathbf{A}$ is atmosphere light of the scene. Besides, $\mathds{1}$ is a matrix whose pixel values are 1 and $\odot$ indicates element-wise multiplication. The transmission map $\mathbf{T}$ encodes influence from vapors to generate rain images. Based on this model, deraining methods propose various CNNs to predict $\mathbf{T}$, $\mathbf{A}$, and $\mathbf{S}$ to calculate background image $\mathbf{J}$.

Rain streaks and vapors are entangled with each other in practice. Removing them separately is not feasible~\cite{li-cvpr19-heavy}. Meanwhile, this entanglement makes Eq.~\eqref{eq:rain-py1} difficult to explicitly model both. The limitation raises that the transmission map and rain streaks are not estimated well. The incorrect estimation brings unnatural illumination and color contrasts on the background. Although a generative adversarial network~\cite{li-cvpr19-heavy} is employed to refine background beyond model constraint, the illumination and color contrasts are not completely corrected as shown in Fig.~\ref{fig:intro} (d).

In this work, we rethink rain image formation by delving rainy model itself. We observe that in Eq.~\eqref{eq:rain-py1}, both rain streaks and background are modeled to have the same properties. This is due to the meaning which two terms convey in Eq.~\eqref{eq:rain-py1}. The first term $\mathbf{T} \odot (\mathbf{J} + \sum^{n}_{i=1} \mathbf{S}_{i})$ indicates that both $\mathbf{S}_{i}$ and $\mathbf{J}$ are transmitted via $\mathbf{T}$. The rain streaks are regarded as part of the background to be transmitted. The second term $(\mathds{1}-\mathbf{T}) \odot \mathbf{A}$ shows that rain streaks do not contribute to atmosphere light transmission because only vapors are considered in $\mathbf{T}$. As rain streaks and vapors are entangled with each other, the modeling of rain streaks as background is not accurate. Based on this observation, we propose a visual model which formulates rain streaks as transmission medium. The entanglement of rain streaks and vapors is modeled properly from the transmission medium perspective. We show the proposed model in the following:
\begin{equation}\label{eq:rain-py2}
  \mathbf{I} = (\mathbf{T_s}+\mathbf{T_v})\odot \mathbf{J} + [\mathds{1}-(\mathbf{T_s}+\mathbf{T_v})]\odot \mathbf{A}
\end{equation}
where $\mathbf{T_s}$ and $\mathbf{T_v}$ are the transmission map of rain streaks and vapors, respectively. In our model, all the variables are extended to the same size, so that we utilize element-wise multiplication to describe the relationship of variables.

Rain streaks appear in various shapes and directions. This phenomenon is more obvious in heavy rain. In order to effectively predict $\mathbf{T_s}$, we propose an encoder-decoder CNN with ShuffleNet units~\cite{zhang-cvpr18-shufflenet} named SNet. The group convolutions and channel shuffle improve network robustness upon diverse rain streaks. The learned multiple groups in ShuffleNet units are able to capture anisotropic appearances of rain streaks. Furthermore, we predict transmission map of vapors (i.e., $\mathbf{T_v}$) by using a VNet where there is a spatial pyramid pooling (SPP) structure. VNet takes the concatenation of $\mathbf{I}$ and $\mathbf{T_s}$ as input and use SPP to capture its global and local features in multi-scales for compact representation. On the other hand, we propose an encoder CNN named ANet to predict atmosphere light $\mathbf{A}$. ANet is pretrained by using training data in a simplified low transmission condition, under which we obtain estimated labels of $\mathbf{A}$ from rainy image $\mathbf{I}$. After pretraining ANet, we jointly train SNet, VNet and ANet by measuring the difference between the calculated background $\mathbf{J}$ and the ground truth background. The learned networks well predict $\mathbf{T_s}$, $\mathbf{T_v}$, and $\mathbf{A}$, which are further transformed to generate background images. We evaluate the proposed method on standard benchmark datasets. The proposed visual model is shown effective to model transmission maps of rain streaks and vapors, which are removed in the generated background images.

We summarize the contributions of this work as follows:
\begin{itemize}[noitemsep,nolistsep]
  \item We remodel the rain image formation by formulating rain streaks as transmission medium. The rain streaks and vapor contribute together to transmit both scene content and atmosphere light into input rain images.
  \item We propose SNet, VNet and ANet to learn rain streaks transmission map, vapor transmission map and atmosphere light. These three CNNs are jointly trained to facilitate the rain image restoration process.
  \item Experiments on the benchmark datasets show the proposed model is effective to predict rain streaks and vapors. The proposed deraining method performs favorably against state-of-the-art approaches.
\end{itemize}

\section{Related Work}

Single image deraining originate from dictionary learning \cite{mairal-jmlr10-online} to solve the negative impact of various rain streaks on the background \cite{wang-tip17-a,wang-icip16-a,kang-tip12-automatic,chen-tcsvt14-visual,zhang-icme06-rain,huang-icme12-context,huang-tmm14-tmm,chen-iccv13-a,luo-iccv15-removing,li-cvpr16-rain}. Recently, deep learning has obtained better deraining performances compared with the conventional methods. Prevalent deep learning based deraining methods can be categorized as direct mapping method, residual based method and scattering model based methods. Direct mapping based methods directly estimate rain-free background from the observed rainy images via novel CNN networks. It includes the work \cite{wang-cvpr19-spatial}, in which a dataset is first built by incorporating temporal priors and human supervision. Then, a novel SPANet is proposed to solve the random distribution of rain streaks in a local-to-global manner.

A residual rain model is proposed in residual based methods to formulate a rainy image as a summation of the background layer and rain layers. It covers majority of existing deraining methods. For example, Fu et al. train their DerainNet in high-frequency domain instead of the image domain to extract image details to improve deraining visual quality \cite{fu-tip17-clearing}. In the meantime, inspired by the deep residual network (ResNet) \cite{he-cvpr15-deep}, a deep detail network which is also trained in high-pass domain was proposed to reduce the mapping range from input to output, to make the learning process easier \cite{fu-cvpr17-removing}. Yang et al. create a new model which introduces atmospheric light and transmission to model various rain streaks and veiling effect, but the rainy image is finally decomposed into a rain layer and background layer by their JORDER network. During the training, a binary map is learnt to locate rain streaks to guide the deraining network \cite{yang-cvpr17-deep,yang-pami19-joint}. In \cite{zhang-cvpr18-density}, the density of rain streaks is classified into three classes and automatically estimated to guide the training of a multi-stream densely connected DID-MDN structure which can better characterize rain streaks with various shape and size. Li et al. regard the rain in rainy images as a summation of multiple rain streak layers, then use a recurrent neural network to remove rain streaks state-wisely \cite{li-eccv18-recurrent}. Hu et al. study the visual effect of rain to scene depth, based on which fog is introduced to model the formation of rainy images and the depth feature is learned to guide their end-to-end network to obtain rain layer \cite{hu-cvpr19-depth}. In \cite{ren-cvpr19-progressive}, a better and simpler deraining baseline is proposed by considering the network structure, input and output of network, and the loss functions.

In the scattering model based methods, atmospheric light and transmission of vapor are rendered and learned to remove rain streaks as well as vapor effect, but rain streaks are treated the same as the background rather than the transmission medium \cite{li-cvpr19-heavy}. Different from existing approaches, we reformulate rainy image generation by modeling rain streaks as transmission medium instead of background content, and use two transmission maps to model the influence of rain streaks and vapor on the background. This formulation naturally models the entanglement of rain streaks and vapors and produce more robust results.

\section{Proposed Algorithm}
We show an overview of the pipeline in Fig. \ref{fig:pipeline}. It consists of SNet, VNet and ANet to estimate transmission maps and atmosphere light. The background image can then be computed as follows:
\begin{equation}\label{eq:background}
  \mathbf{J} = \{\mathbf{I}-[\mathds{1}-(\mathbf{T_s}+\mathbf{T_v})]\odot \mathbf{A}\} \oslash (\mathbf{T_s}+\mathbf{T_v}),
\end{equation}
where $\oslash$ is the element-wise division operation. In the following, we first illustrate the network structure of SNet, VNet, and ANet, then we show how we train these three networks in practice and elucidate how these networks function in rain image restoration.

\renewcommand{\tabcolsep}{1pt}
\def\swone{1\linewidth}
\begin{figure*}[t]
\begin{center}
\begin{tabular}{c}
\includegraphics[width=\swone]{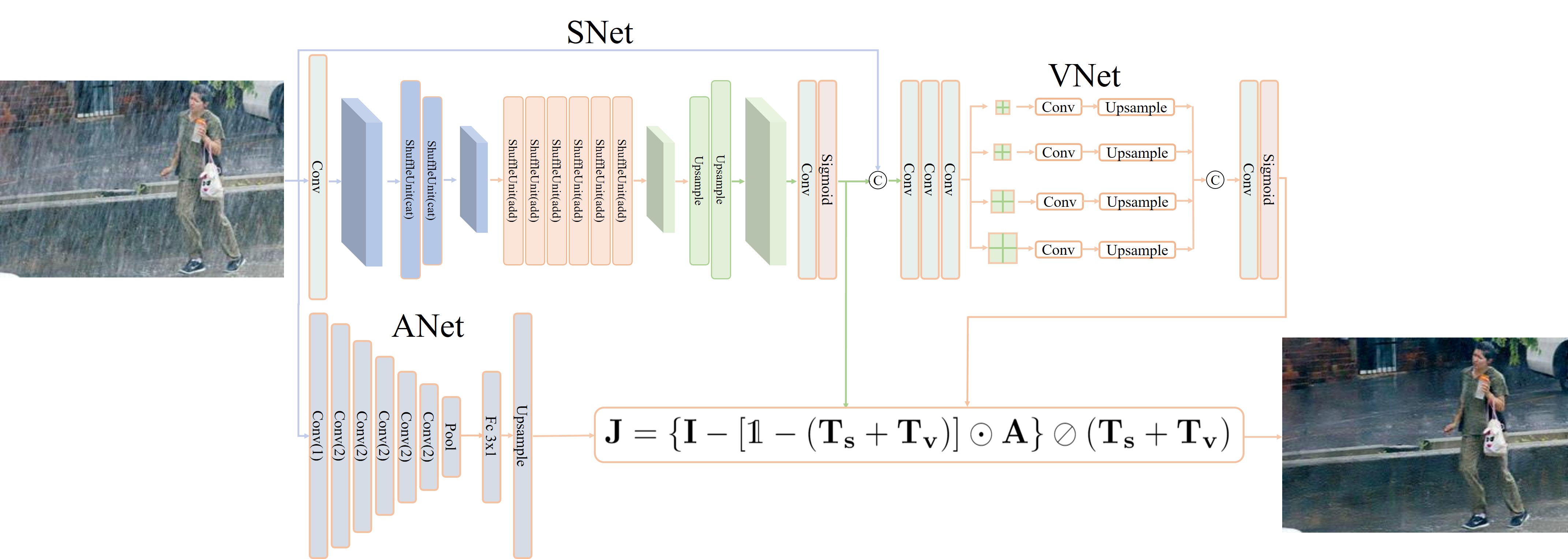}\\
\end{tabular}
\end{center}
\caption{This figure shows our network structure. The pool denotes adaptive average pooling operation. The Upsample operation after triangle-shaped network extends the atmospheric light $\mathbf{A}$ to the image size. The notations $\odot$ and $\oslash$ are the pixel-wise multiplication and division, respectively}
\label{fig:pipeline}
\end{figure*}

\renewcommand{\tabcolsep}{1pt}
\def\swone{0.5\linewidth}
\begin{figure}[t]
\begin{center}
\begin{tabular}{c}
\includegraphics[width=\swone]{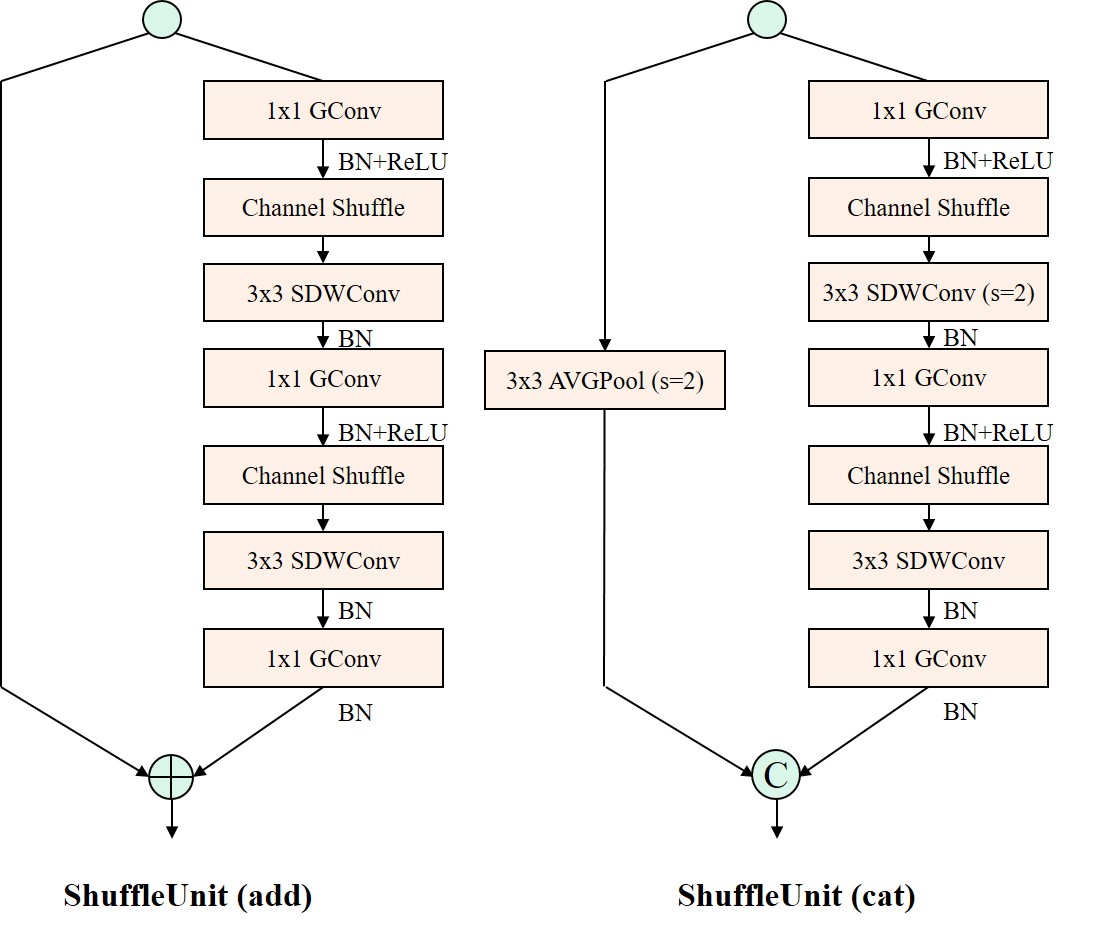}\\
\end{tabular}
\end{center}
\caption{This figure shows our revised ShuffleNet Units. ShuffleUnit(add) can keep image size unchanged, and ShuffleUnit(cat) downsamples image once. $+$ and C mean addition and concatenation respectively.}
\label{fig:shuffle}
\end{figure}

\subsection{SNet}\label{sec:snet}

We propose a SNet that takes rain image as input and predicts rain streak transmission maps $\mathbf{T_s}$. SNet is an encoder-decoder CNN with ShuffleNet units that consist of group convolutions and shuffling operations. The input CNN features are partially captured by different groups and then shuffled to fuse together. We extend ShuffleNet unit to capture anisotropic representation of rain streaks as shown in Fig. \ref{fig:pipeline}. Our extension is shown in Fig.~\ref{fig:shuffle} where we increase the number of group convolutions and deep separable convolution.
The features of different groups in single unit will be more discriminative by twice grouping to boost the global feature grouping. Moreover, the depthwise convolution is symmetrically padded (SDWConv) to decrease the influence of padded $0$ on the image edges.
Finally, we upsample the feature map to original size and convolution layers are followed to fuse multi-group features. The prediction of $\mathbf{T_s}$ on SNet can be written as:
\begin{equation}\label{eq:SNet}
  \mathbf{T_s}=\mathcal{S}(\mathbf{I})
\end{equation}
where $\mathbf{I}$ is the rain image and $\mathcal{S}(\cdot)$ is the SNet inference.

\subsection{VNet}
We propose a VNet that captures multi-scale features to predict vapor transmission maps $\mathbf{T_v}$. VNet takes the concatenation of rain image $\mathbf{I}$ and $\mathbf{T_s}$ as input where $\mathbf{T_s}$ provides the global intensity information for $\mathbf{T_v}$ and $\mathbf{I}$ supplies the local background information, as different local areas have different vapor intensity. Compared with anisotropic rain streaks, vapor is locally homogeneous and the values of different areas have high correlation. VNet utilizes SPP structure to capture global and local features to provide compact feature representation for $\mathbf{T_v}$ as shown in Fig. \ref{fig:pipeline}. The prediction of $\mathbf{T_v}$ on VNet can be written as:
\begin{equation}\label{eq:VNet}
  \mathbf{T_v}=\mathcal{V}(\operatorname{cat}(\mathbf{I}, \mathbf{T_s})).
\end{equation}

\subsection{ANet}

We propose an encoder network ANet to predict atmosphere light. Its structure is shown in Fig.~\ref{fig:pipeline}. The network inference can be written as:
\begin{equation}\label{eq:ANet}
    \mathbf{A} = \mathcal{A}(\mathbf{I}),
\end{equation}
where $\mathcal{A}(\cdot)$ is the ANet inference. As atmosphere light is usually considered constant in the rain image, the output of the encoder is a $3\times 1$ vector. We use an ideal form of our rain model Eq. \eqref{eq:rain-py2} to create labels of atmospheric light from rain images to pretrain ANet. Then, we integrate ANet into the whole pipeline for joint training. The details are presented in the following.

\subsection{Network Training}
\label{sec:nettrain}

The pipeline of the whole network consists of a SNet, a VNet, and an ANet to predict $\mathbf{T_s}$, $\mathbf{T_v}$, and $\mathbf{A}$, respectively. Then, we will generate $\mathbf{J}$ according to Eq. \eqref{eq:background}. We first pretrain ANet using labels from a simplified condition and perform joint training of these three networks.

\begin{algorithm}[t]
\renewcommand{\algorithmicrequire}{\textbf{Input:}}
\caption{Pretraining ANet}
\label{alg:train_ANet}
\begin{algorithmic}[1]
\REQUIRE Rainy images $\{ \mathbf{I}^{\{t\}} \}$.
\FOR{$i=1$ to epoch}
\FOR{$j=1$ to batchnum}
\STATE Locate rain pixels for $\mathbf{I}^{\{t\}}$ via \cite{yang-cvpr17-deep}.
\STATE Based on Eq. \eqref{eq:real}, find highest rainy pixel as the ground truth atmospheric light $\mathbf{A}^{\{t \}}$.
\STATE Calculate $\mathcal{A}(\mathbf{I}^{\{ t\}})$ via Eq. \eqref{eq:ANet}.
\STATE Updating $\mathcal{A}(\cdot)$ via loss \eqref{eq:lossANet}.
\ENDFOR
\ENDFOR
\renewcommand{\algorithmicensure}{\textbf{Output:}}
\ENSURE Learned atmospheric light $\mathbf{A}$.
\end{algorithmic}
\end{algorithm}

\subsubsection{Pretraining ANet.}
\label{sec:pretrainanet}
Sample collection is crucial for pretraining ANet as the ground truth value of atmosphere light is difficult to obtain. Instead of empirically modeling $\mathbf{A}$ as a uniform distribution~\cite{li-cvpr19-heavy}, we generate labels under a simplified condition based on our rain model Eq. \eqref{eq:rain-py2}, where the transmission maps of both rain streaks and vapors are 0. For one pixel $x$ in rain image $\mathbf{I}$, $\mathbf{T_s}(x)+\mathbf{T_v(x)}=0$, our visual model of rain image formation can be written as:
\begin{equation}\label{eq:ideal}
    \mathbf{I}(x)=\mathbf{A}(x)
\end{equation}
where the pixel value of atmosphere light is equal to that of rain image. In practice, the values of transmission at rain pixel $x$ with high intensity approach 0 (i.e., $\mathbf{T_s}(x)+\mathbf{T_v}(x) \approx 0$). Our model in Eq.~\ref{eq:rain-py2} can be approximated by:
\begin{equation}\label{eq:real}
    \mathbf{I}(x)=[1-(\mathbf{T_s}(x)+\mathbf{T_v}(x))]*\mathbf{A}(x)
\end{equation}
where the maximum value of $\mathbf{I}(x)$ at rain streak pixels is $\mathbf{A}(x)$. We use \cite{yang-cvpr17-deep} to detect rainy pixels in $\mathbf{I}$ and identify the maximum intensity value as ground truth atmospheric light $\mathbf{A}$. In this simplified form, we obtain labels for ANet and train it using the following form:
\begin{equation}\label{eq:lossANet}
  \mathcal{L}_\mathbf{A}=\frac{1}{N}\sum_{t=1}^{N}||\mathcal{A}(\mathbf{I}_t)-\mathbf{A}_t||^2
\end{equation}
where $N$ is the number of training samples. The  algorithm is in Algorithm \ref{alg:train_ANet}.

\subsubsection{Pretraining SNet.}
We pretrain SNet by assuming an ideal case where vapors do not contribute to the transmission (i.e., $\mathbf{T_v}=0$). We use the input rain image $\mathbf{I}$ and ground truth restoration image $\mathbf{J}$ to train SNet. The objective function can be written as follows:
\begin{equation}\label{eq:lossSNet}
  \mathcal{L}_\mathbf{S}=\frac{1}{N}\sum_{t=1}^{N}||\mathcal{J}(\mathbf{I}_{t})-\mathbf{J}_{t}||^2_{F}
\end{equation}
where $\mathcal{J}(\mathbf{I}_t)=\{\mathbf{I}_{t}-[\mathds{1}-\mathcal{S}(\mathbf{I}_t)]\odot\mathcal{A}(\mathbf{I}_t)\}\oslash\mathcal{S}(\mathbf{I}_t)$ derives from Eq.~\ref{eq:background}. More details are shown in Algorithm \ref{alg:train_SNet}.

\subsubsection{Joint Training.}
After pretraining ANet and SNet, we perform joint training of the whole network. The overall objective function can be written as:
\begin{eqnarray}\label{eq:loss-VNet}
  \nonumber \mathcal{L}_{\rm total}=\lambda_1\cdot \frac{1}{N}\sum_{t=1}^{N}||\triangledown\mathcal{J}(\mathbf{I}_t)-\triangledown \mathbf{J}_t||^2_F
  +\lambda_2\cdot \frac{1}{N}\sum_{t=1}^{N} ||\mathcal{J}(\mathbf{I}_t)-\mathbf{J}_t||_{1}
\end{eqnarray}
where $\triangledown$ is the gradient operator in both horizontal and vertical directions, $\lambda_1$ and $\lambda_2$ are constant weights.
The value $\mathcal{J}(\mathbf{I}_t)$ is from Eq. \eqref{eq:background} consisting of $\mathbf{T_s}$, $\mathbf{T_v}$, and $\mathbf{A}$. These variables are predicted from SNet, VNet, and ANet, respectively. We perform joint training to these networks. As VNet takes the concatenation of $\mathbf{I}$ and $\mathbf{T_s}$ as input, we back propagate the network gradient to SNet via VNet. The details of our joint training is shown in Algorithm \ref{alg:jonit_train}.

\begin{algorithm}[t]
\renewcommand{\algorithmicrequire}{\textbf{Input:}}
\caption{Pretraining SNet}
\label{alg:train_SNet}
\begin{algorithmic}[1]
\REQUIRE Rainy images $\{ \mathbf{I}^{\{t\}} \}$ and ground truth background $\{ \mathbf{J}^{\{t\}} \}$.
\renewcommand{\algorithmicrequire}{\textbf{Initialization:}}
\REQUIRE $\mathbf{T_v}=0$ in Eq. \eqref{eq:background}, $\mathcal{A}(\cdot)$ is initialized with pretrained model in Alg. \ref{alg:train_ANet}.
\FOR{$i=1$ to epoch}
\FOR{$j=1$ to batchnum}
\STATE Calculate $\mathcal{A}(\mathbf{I}^{\{ t\}})$ for $\{ \mathbf{I}^{\{t\}} \}$ via Eq. \eqref{eq:ANet}.
\STATE Calculate $\mathcal{S}(\mathbf{I}^{\{ t\}})$ via Eq. \eqref{eq:SNet}.
\STATE Calculate $\mathcal{J}(\mathbf{I}^{\{ t\}})$ via Eq. \eqref{eq:background}.
\STATE Updating $\mathcal{S}(\cdot)$ and fine tuning $\mathcal{A}(\cdot)$ via loss \eqref{eq:lossSNet}.
\ENDFOR
\ENDFOR
\renewcommand{\algorithmicensure}{\textbf{Output:}}
\ENSURE Learned transmission map $\mathbf{T_s}$ of rain streaks.
\end{algorithmic}
\end{algorithm}

\begin{algorithm}[t]
\renewcommand{\algorithmicrequire}{\textbf{Input:}}
\caption{Joint training}
\label{alg:jonit_train}
\begin{algorithmic}[1]
\REQUIRE Rainy images $\{ \mathbf{I}^{\{t\}} \}$ and ground truth background $\{ \mathbf{J}^{\{t\}} \}$.
\renewcommand{\algorithmicrequire}{\textbf{Initialization:}}
\REQUIRE $\mathcal{A}(\cdot)$ is initialized with fine tuned model in Alg. \ref{alg:train_SNet}, $\mathcal{S}(\cdot)$ is initialized with pretrained model in Alg. \ref{alg:train_SNet}.
\FOR{$i=1$ to epoch}
\FOR{$j=1$ to batchnum}
\STATE Calculate $\mathcal{A}(\mathbf{I}^{\{ t\}})$ for $\{ \mathbf{I}^{\{t\}} \}$ via Eq. \eqref{eq:ANet}.
\STATE Calculate $\mathcal{S}(\mathbf{I}^{\{ t\}})$ via Eq. \eqref{eq:SNet}.
\STATE Calculate $\mathcal{V}(\operatorname{cat}(\mathbf{I}^{\{ t\}}, \mathcal{S}(\mathbf{I}^{\{ t\}})))$ via Eq. \eqref{eq:VNet}.
\STATE Calculate $\mathcal{J}(\mathbf{I}^{\{ t\}})$ via Eq. \eqref{eq:background}.
\STATE Updating $\mathcal{V}(\cdot)$ and fine tuning $\mathcal{A}(\cdot)$ and $\mathcal{S}(\cdot)$ via loss \eqref{eq:loss-VNet}.
\ENDFOR
\ENDFOR
\renewcommand{\algorithmicensure}{\textbf{Output:}}
\ENSURE Learned transmission map $\mathbf{T_s}$ of rain streaks.
\end{algorithmic}
\end{algorithm}

\renewcommand{\tabcolsep}{1pt}
\def\swfour{0.24\linewidth}
\begin{figure}[t]
\begin{center}
\begin{tabular}{cccc}
\includegraphics[width=\swfour]{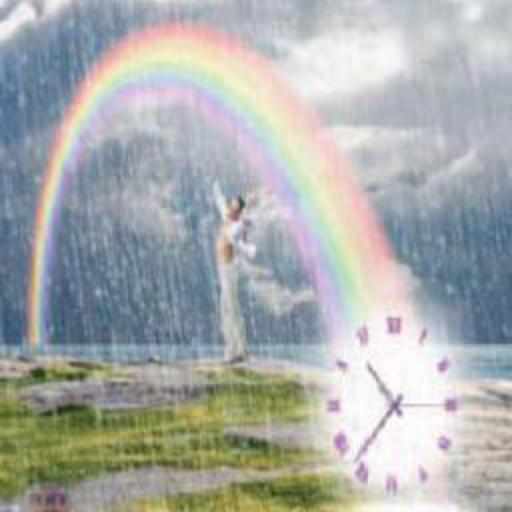}&\includegraphics[width=\swfour]{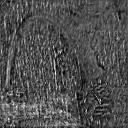}&\includegraphics[width=\swfour]{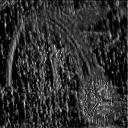}&\includegraphics[width=\swfour]{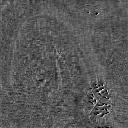}\\
\includegraphics[width=\swfour]{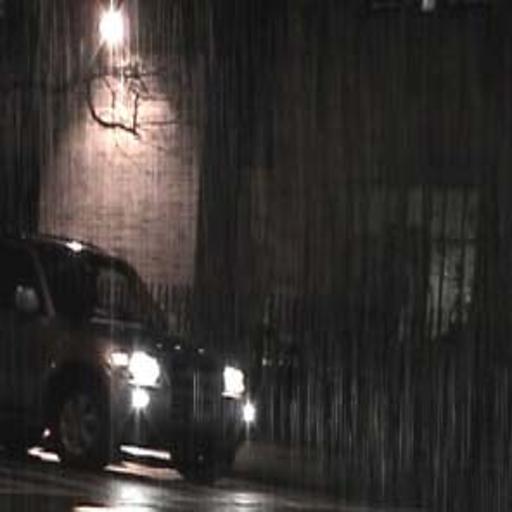}&\includegraphics[width=\swfour]{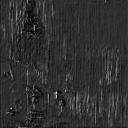}&\includegraphics[width=\swfour]{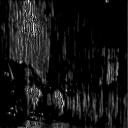}&\includegraphics[width=\swfour]{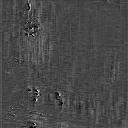}\\
(a) & (b) & (c) & (d) \\
\end{tabular}
\end{center}
\caption{Feature maps of different convolution groups. (a) Input rainy images. (b) Features in 1st group. (c) Features in 2nd group. (d) Features in 3rd group. The features of rain streaks in the first group is always slim, the second group extract rain streaks with relatively large size, and the third group contains features of homogeneous vapor.}
\label{fig:groupfea}
\end{figure}

\renewcommand{\tabcolsep}{1pt}
\def\swseven{0.14\linewidth}
\begin{figure*}[h]
\begin{center}
\begin{tabular}{ccccccc}
\includegraphics[width=\swseven]{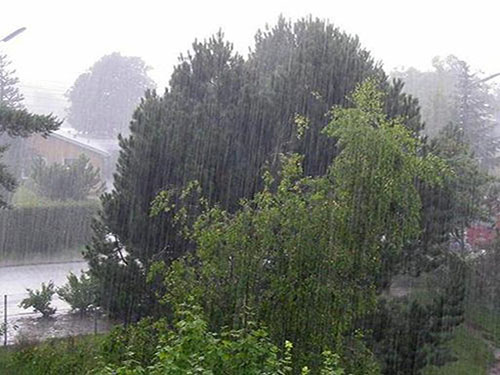}&
\includegraphics[width=\swseven]{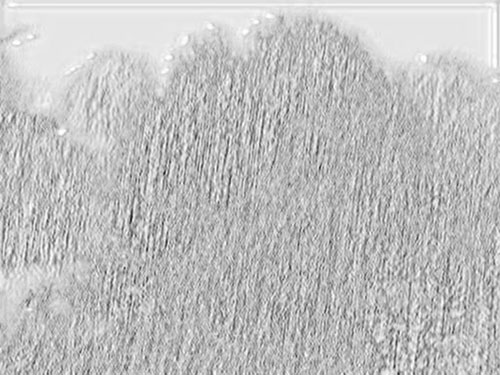}&
\includegraphics[width=\swseven]{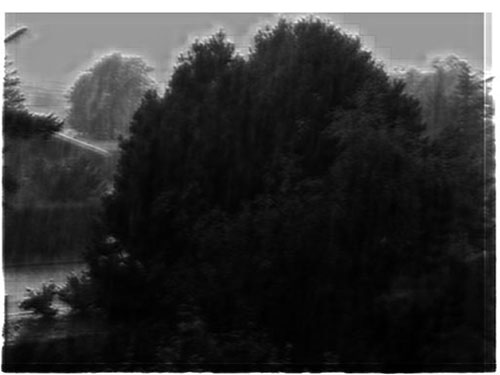}&
\includegraphics[width=\swseven]{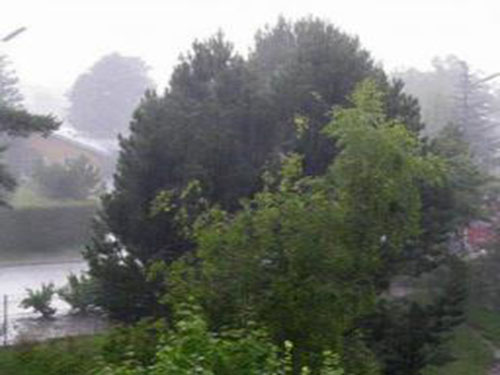}&
\includegraphics[width=\swseven]{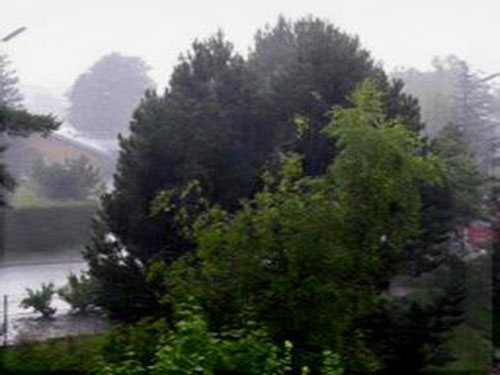}&
\includegraphics[width=\swseven]{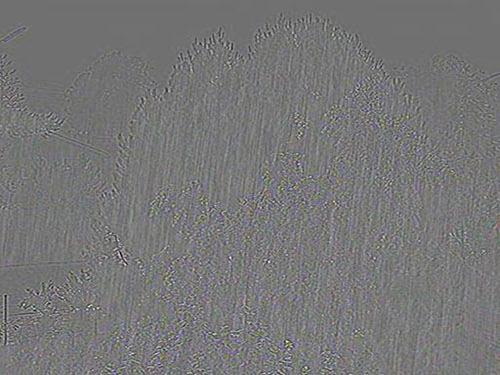}&
\includegraphics[width=\swseven]{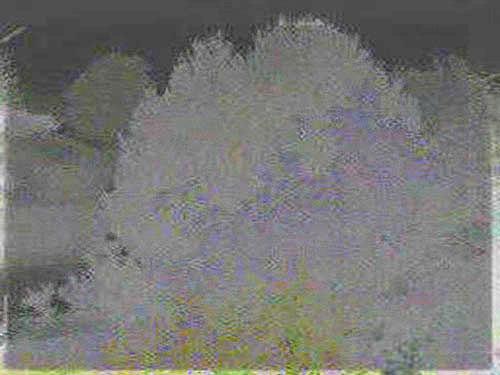}\\
\includegraphics[width=\swseven]{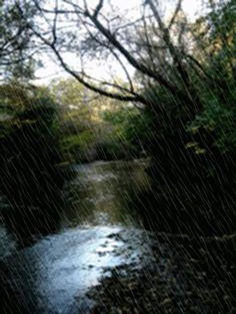}&
\includegraphics[width=\swseven]{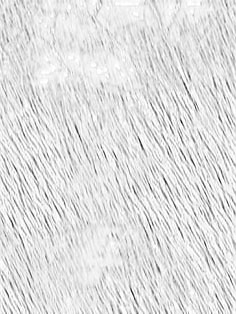}&
\includegraphics[width=\swseven]{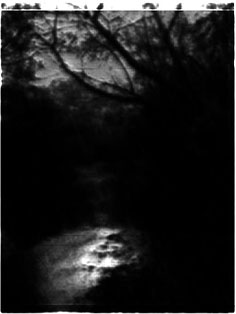}&
\includegraphics[width=\swseven]{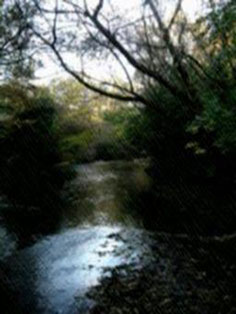}&
\includegraphics[width=\swseven]{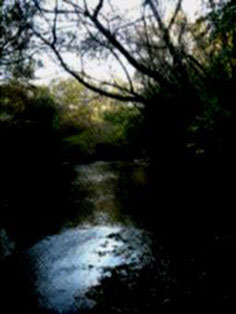}&
\includegraphics[width=\swseven]{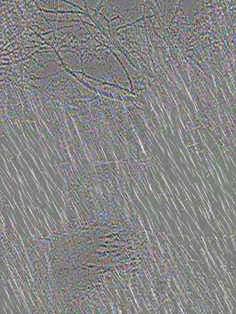}&
\includegraphics[width=\swseven]{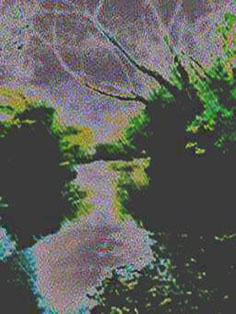}\\
(a) & (b) & (c) & (d) & (e) & (f) & (g) \\
\end{tabular}
\end{center}
\caption{Transmission map of rain streaks and vapor.  Input rainy images are shown in (a). Transmission maps $\mathbf{T_s}$ of rain streaks are in (b). Transmission maps $\mathbf{T_v}$ of vapor are in (c). Deraining results with only using $\mathbf{T_s}$ are in (d). Deraining results with $\mathbf{T_v}$ involved are in (e). The Removed rain streaks are shown in (f) and the removed vapors are shown in (g). $\mathbf{T_s}$ is shown to capture anisotropic rain streaks while $\mathbf{T_v}$ models homogeneous vapors.}
\label{fig:transmap}
\end{figure*}

\subsection{Visualizations}

We visualize the intermediate results of our method to verify the effectiveness of our network. In Section \ref{sec:snet}, we extract the features of rainy image by $3$ separate convolution groups. We show the learned feature maps of different convolution groups in Fig. \ref{fig:groupfea}. The (a)-(c) shows that different groups contain different features of rain streaks in various shapes and sizes. The first group extracts slim rain streaks and their shapes are similar, the second group contains wide rain features and the shapes are diversified. The third group captures  homogeneous feature representations resembling vapors.

Our rain model allows for the anisotropic transmission map of rain streaks, the homogeneous transmission map of vapor and the atmospheric light of rainy scenes. In Fig. \ref{fig:transmap}, we display the learned transmission map $\mathbf{T_s}$ of rain streaks, the transmission map $\mathbf{T_v}$ of vapor and the atmospheric light $\mathbf{A}$. We can see that $\mathbf{T_s}$ captures the various rain streak information and contains the anisotropy of rainy scenes. While $\mathbf{T_v}$ models the influence of vapor, it possesses the similar values in local areas and different areas are separated by object contours. $\mathbf{A}$ keeps relatively high values, which reflects the fact that atmospheric light possesses high illumination in rainy scenes.

\section{Experiments}

To assess the performance of our deraining method quantitatively, the commonly used PSNR and SSIM \cite{wang-tip04-image} are used as our metrics. In order to evaluate our deraining network more robustly, we measure the quality of deraining results by calculating their Frechet Inception Distance (FID) \cite{heusel-nips17-gans} to the ground truth background. FID is defined via the deep features extracted by Inception-V3 \cite{szegedy-cvpr16-rethinking}, smaller values of FID indicate more similar deraining results to the ground truth.
For visual quality evaluation, we show some restored results of real-world and synthetic rainy images. Existing methods \cite{li-eccv18-recurrent,yang-pami19-joint,li-cvpr19-heavy,wang-cvpr19-spatial} are selected to make complete comparisons in our paper. The comparisons with another two methods \cite{zhang-cvpr18-density,fu-cvpr17-removing} are provided in the supplementary file. Except for \cite{yang-pami19-joint,li-cvpr19-heavy,zhang-cvpr18-density} which need additional ground truth configuration, these methods are retrained on the same dataset for fair comparisons.

In the training process, we crop $256 \times 256$ patches from the training samples, and Adam \cite{kingma-iclr15-adam} is used to optimize our network. The learning rate for pretraining ANet is $0.001$. While learning $\mathbf{T_s}$, loss $\mathcal{L}_{\mathbf{S}}$ is to train SNet and fine tune ANet in a joint way, the learning rate for SNet is $0.001$ and the learning rate for ANet is $10^{-6}$. Similarly, in the stage of jointly learning $\mathbf{T_v}$, the learning rate for VNet is $0.001$ and the learning rate for SNet and ANet is $10^{-6}$. The hyper-parameters $\lambda_{1}$ and $\lambda_{2}$ in Eq. \eqref{eq:loss-VNet} are $0.01$ and $1$ respectively. Our network is trained on a PC with NVIDIA 1080Ti GPU based on PyTorch framework. The training is converged at the 20-$th$ epoch. Our code will be released publicly.

\renewcommand{\tabcolsep}{10pt}
\begin{table}[t]
\centering
\caption{PSNR and SSIM of our ablation studies}
\begin{tabular}{ccccc}
\hline
\hline
Datasets & \multicolumn{2}{c}{Rain-I} & \multicolumn{2}{c}{Rain-II} \\  \hline
Metric &    PSNR       &    SSIM       &    PSNR       &    SSIM       \\ \hline
$C_1$ &   27.15        &   0.772        &    25.48       &   0.793        \\
$C_2$ &   27.49        &    0.806       &    28.57       &     0.844      \\
$C_3$ &   31.30        &    0.897       &    33.86       &    0.930       \\ \hline
Ours &    \textbf{31.34}       &    \textbf{0.908}       &    \textbf{34.42}       &    \textbf{ 0.938}      \\ \hline
\hline
\end{tabular}
\label{tab:abla}
\end{table}

\renewcommand{\tabcolsep}{1pt}
\def\swfive{0.18\linewidth}
\begin{figure*}[t]
\begin{center}
\begin{tabular}{ccccc}
\includegraphics[width=\swfive]{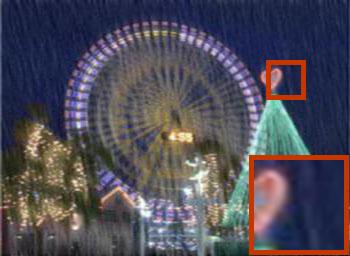}&\includegraphics[width=\swfive]{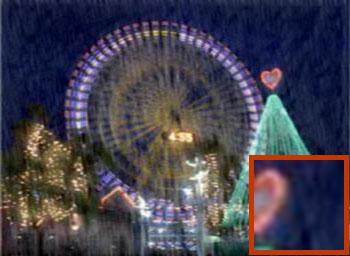}&\includegraphics[width=\swfive]{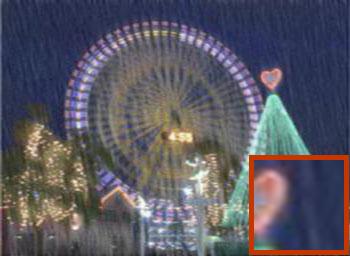}&\includegraphics[width=\swfive]{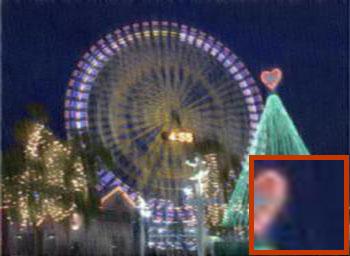}&\includegraphics[width=\swfive]{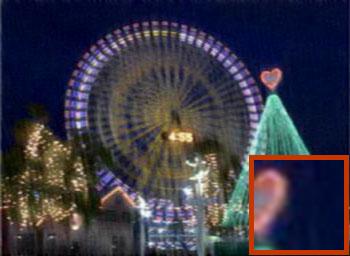}\\
\includegraphics[width=\swfive]{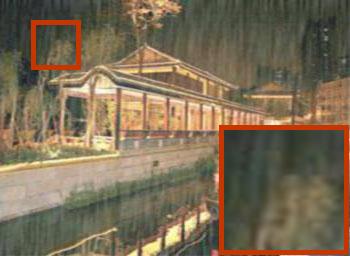}&\includegraphics[width=\swfive]{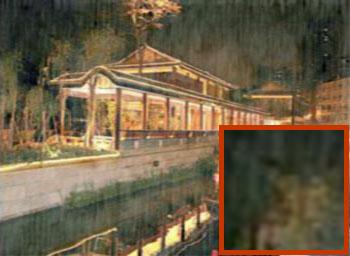}&\includegraphics[width=\swfive]{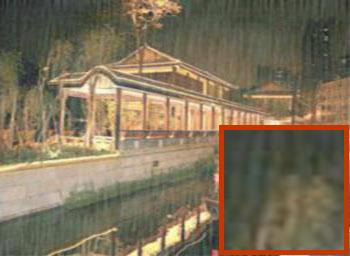}&\includegraphics[width=\swfive]{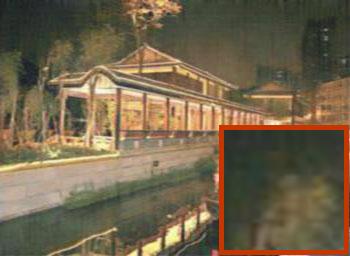}&\includegraphics[width=\swfive]{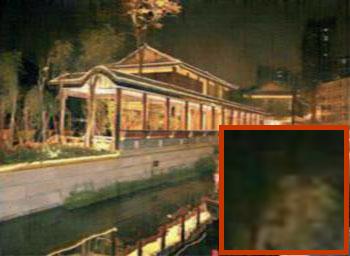}\\
(a) Input & (b) $C_1$ & (c) $C_2$ & (d) $C_3$ & (e) Ours \\
\end{tabular}
\end{center}
\caption{Visual results of ablation studies. (a) Input rainy images. (b)-(e) Deraining results under $C_1$, $C_2$, $C_3$ and the whole pipeline, respectively.}
\label{fig:abla}
\end{figure*}

\renewcommand{\tabcolsep}{5pt}
\begin{table*}[t]
\centering
\caption{PSNR/SSIM comparisons on our three datasets}
\label{tab:quanti}
\begin{tabular}{cccccc}
\hline
\hline
 Methods &  \cite{li-eccv18-recurrent} &  \cite{yang-pami19-joint} & \cite{li-cvpr19-heavy} & \cite{wang-cvpr19-spatial} & Ours \\ \hline
 Rain-I &  27.51/0.897 &  27.69/0.898 & 17.96/0.675 & 28.43/0.848 & \textbf{31.34/0.908} \\
 Rain-II &  26.68/0.830 &  29.97/0.893 & 17.99/0.605 & 30.53/0.905 & \textbf{34.42/0.938} \\
Rain-III &  34.78/0.943 &  28.39/0.902 & 18.48/0.747 & 35.10/0.948 & \textbf{35.91/0.951} \\ \hline
 \hline
\end{tabular}
\end{table*}

\renewcommand{\tabcolsep}{5pt}
\begin{table*}[t]
\centering
\caption{FID comparisons on our three datasets}
\label{tab:quanti-fid}
\begin{tabular}{cccccc}
\hline
\hline
 Methods &  \cite{li-eccv18-recurrent} &  \cite{yang-pami19-joint} & \cite{li-cvpr19-heavy} & \cite{wang-cvpr19-spatial} & Ours \\ \hline
 Rain-I &  62.71 &  101.74 & 104.08 & 81.54 & \textbf{50.66} \\
 Rain-II &  97.30 &  134.54 & 118.10 & 88.15 & \textbf{67.18} \\
Rain-III &  81.42 &  89.63 & 134.34 & 80.68 & \textbf{79.86} \\ \hline
 \hline
\end{tabular}
\end{table*}

\renewcommand{\tabcolsep}{1pt}
\def\swseven{0.14\linewidth}
\begin{figure*}[t]
\begin{center}
\begin{tabular}{ccccccc}
\includegraphics[width=\swseven]{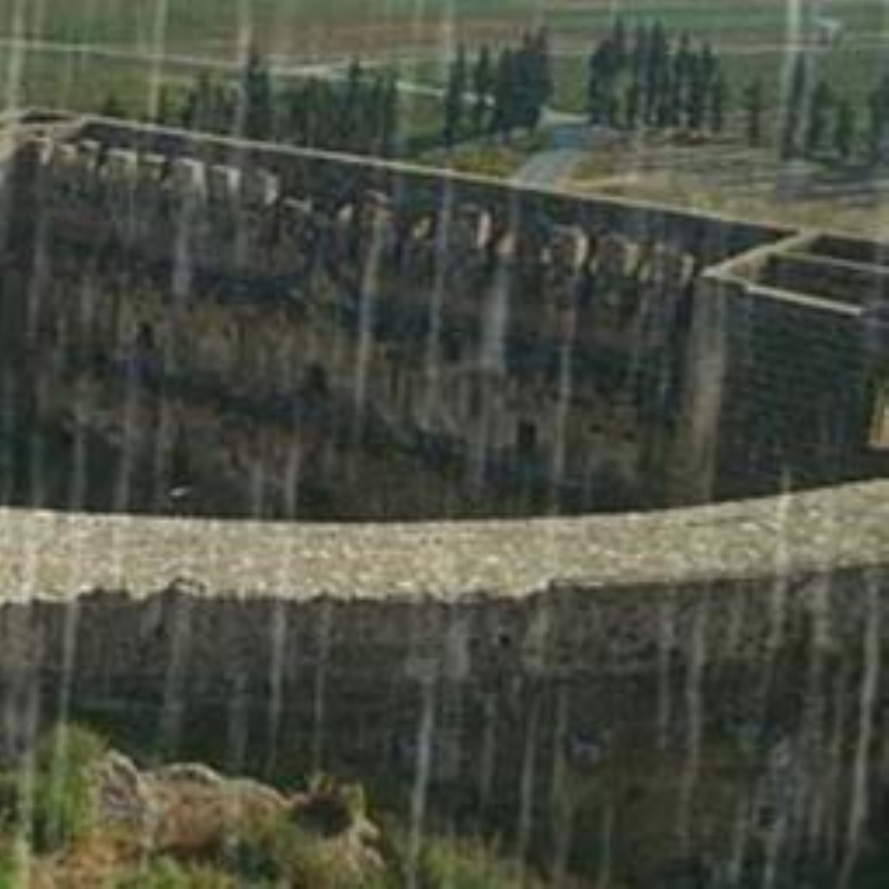}&
\includegraphics[width=\swseven]{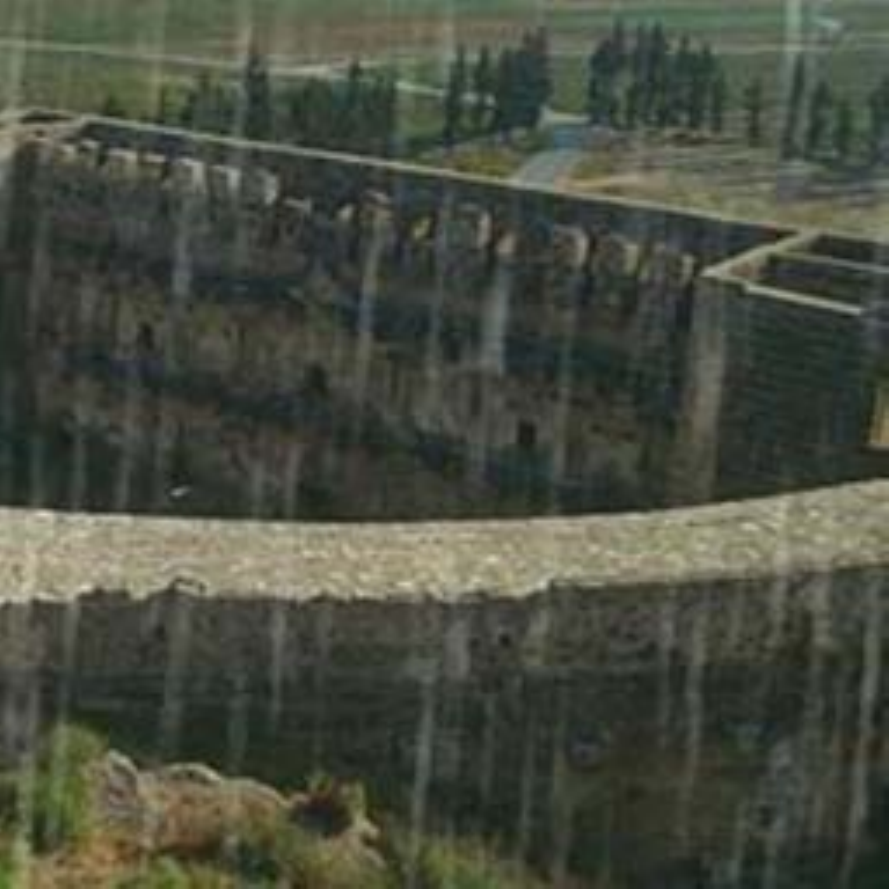}&
\includegraphics[width=\swseven]{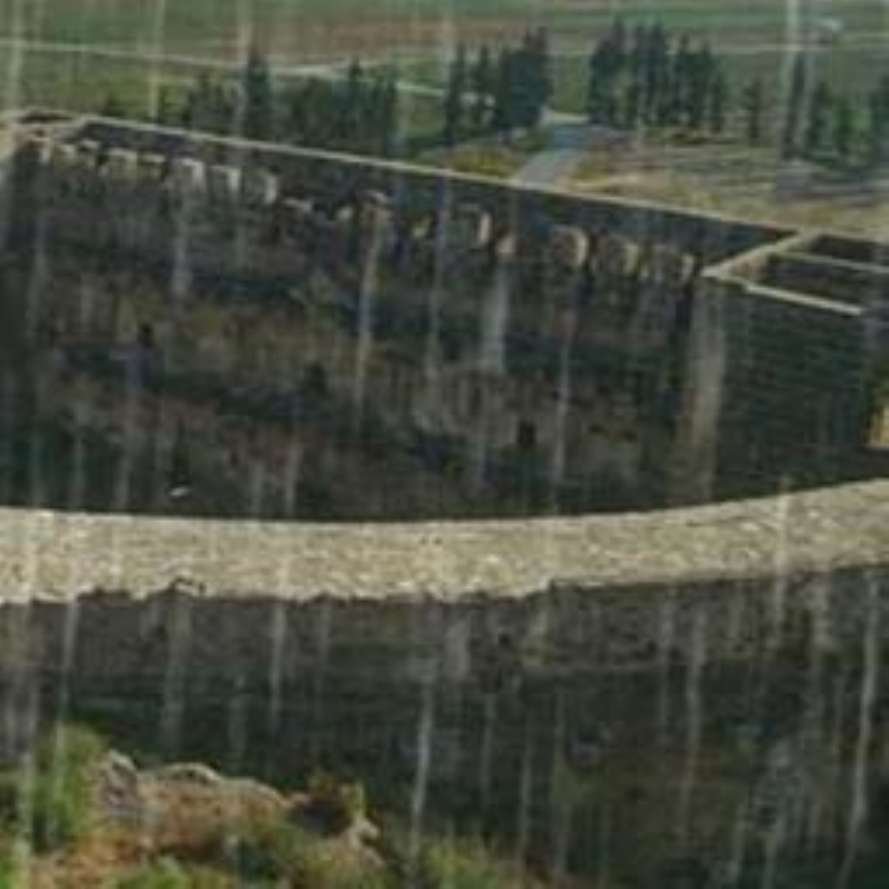}&
\includegraphics[width=\swseven]{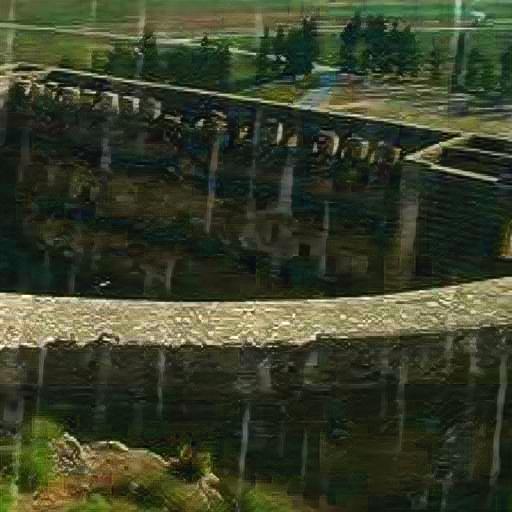}&
\includegraphics[width=\swseven]{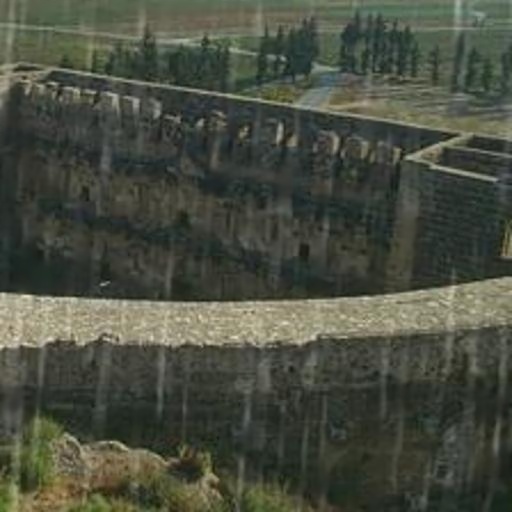}&
\includegraphics[width=\swseven]{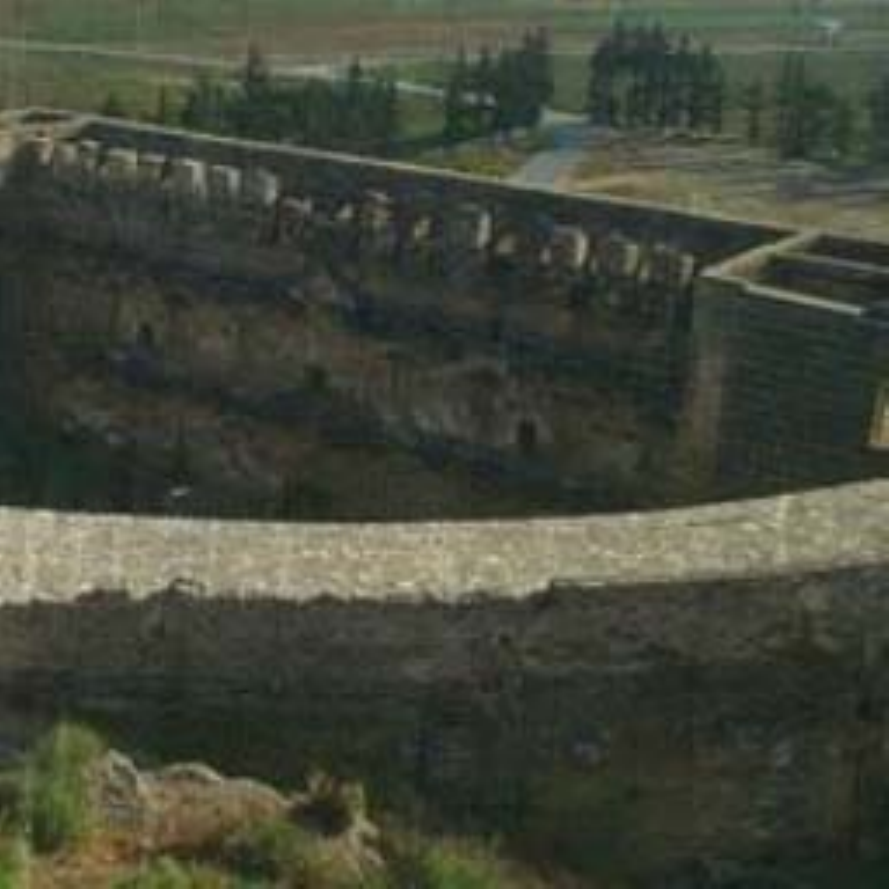}&
\includegraphics[width=\swseven]{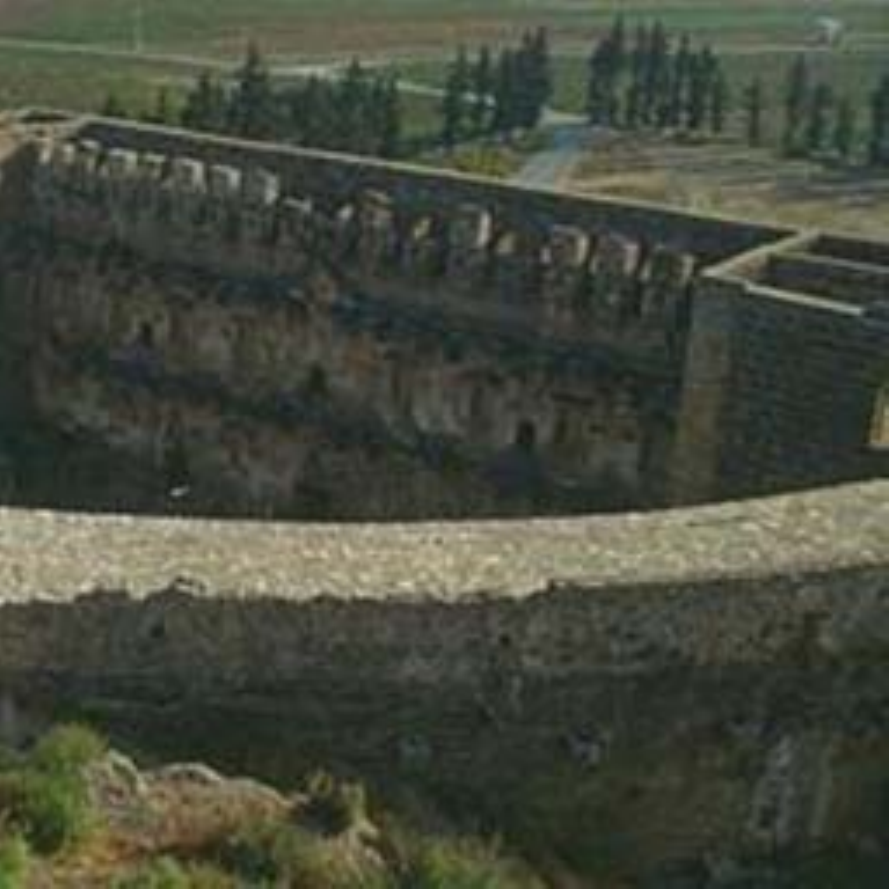}\\
\includegraphics[width=\swseven]{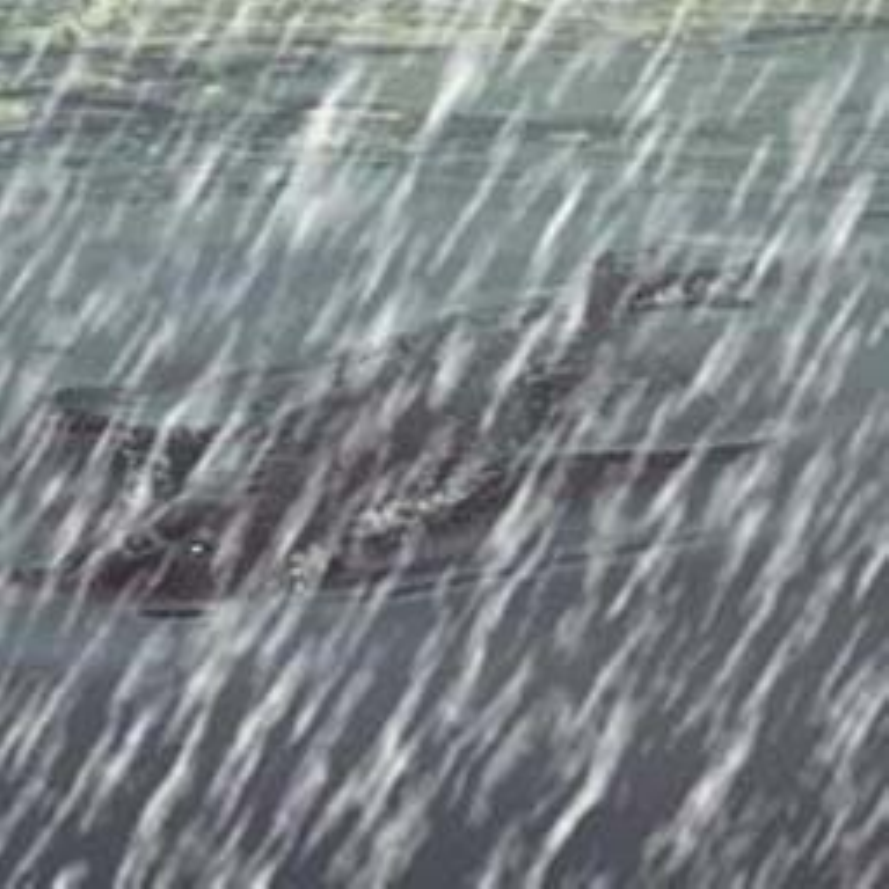}&
\includegraphics[width=\swseven]{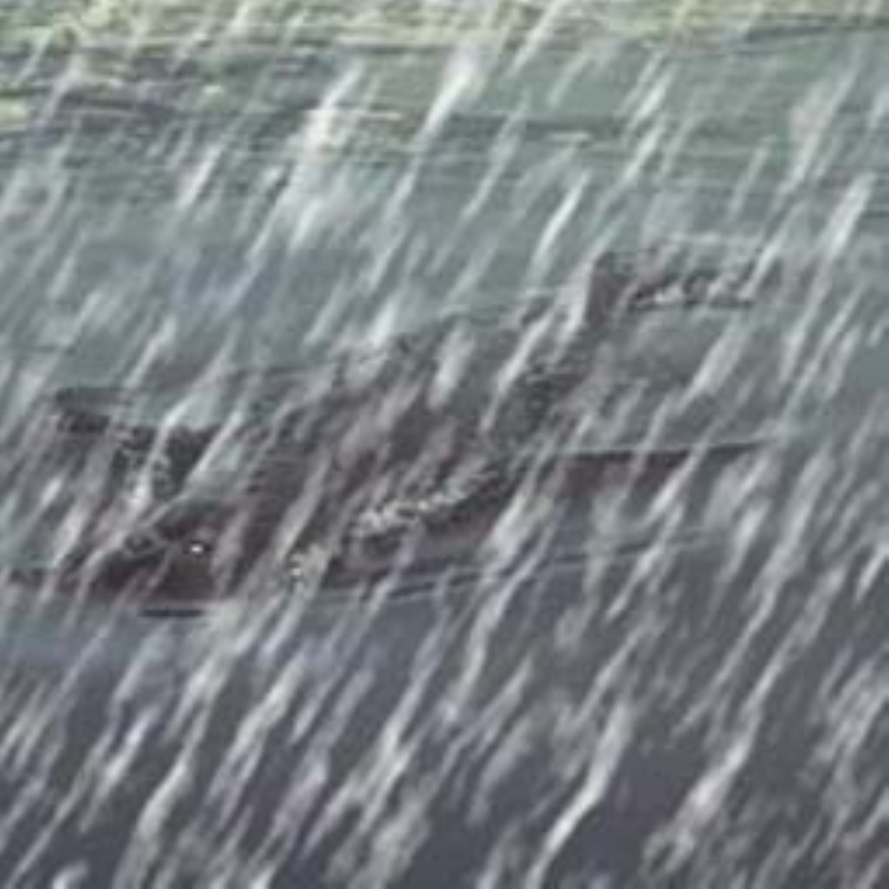}&
\includegraphics[width=\swseven]{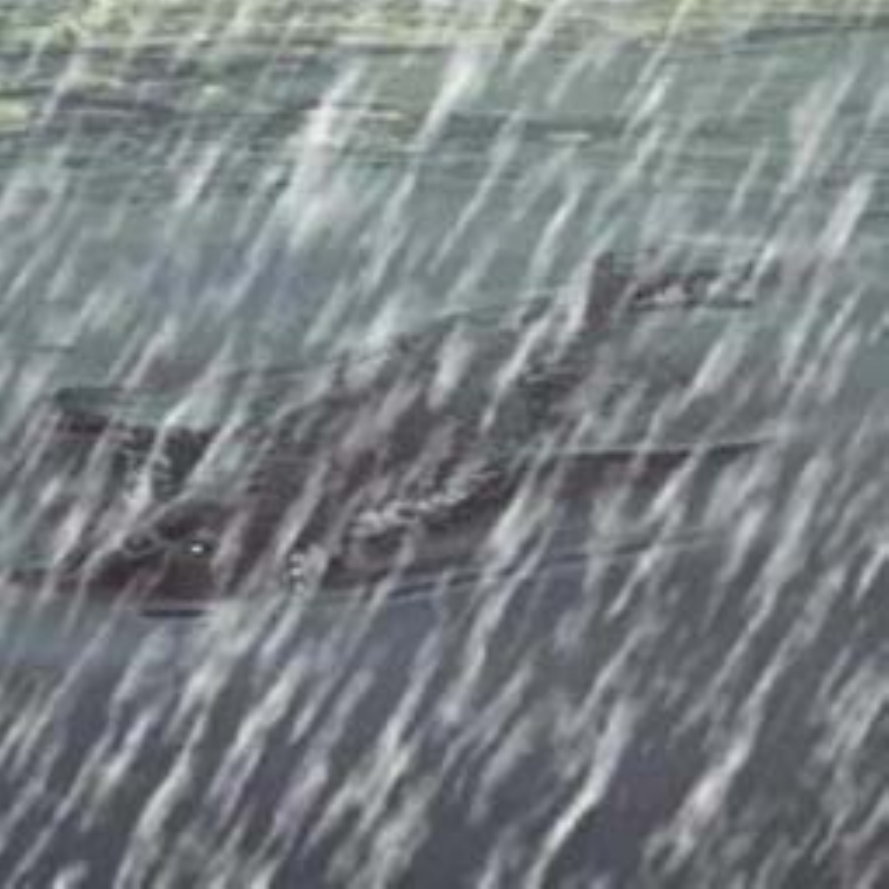}&
\includegraphics[width=\swseven]{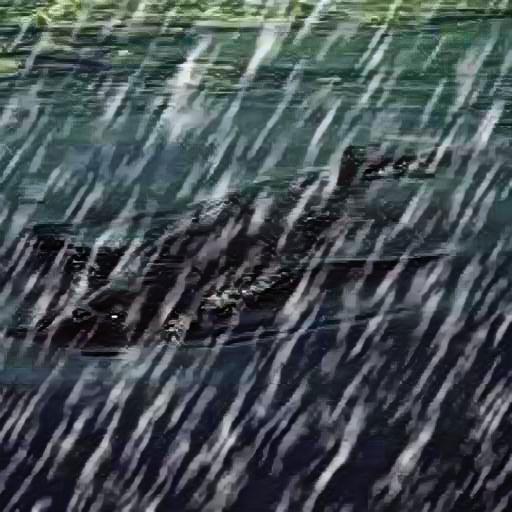}&
\includegraphics[width=\swseven]{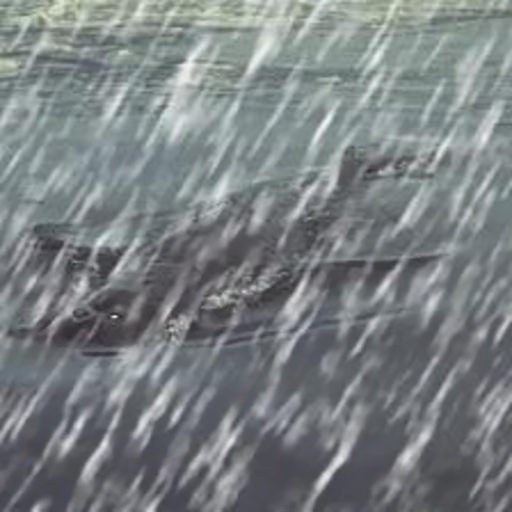}&
\includegraphics[width=\swseven]{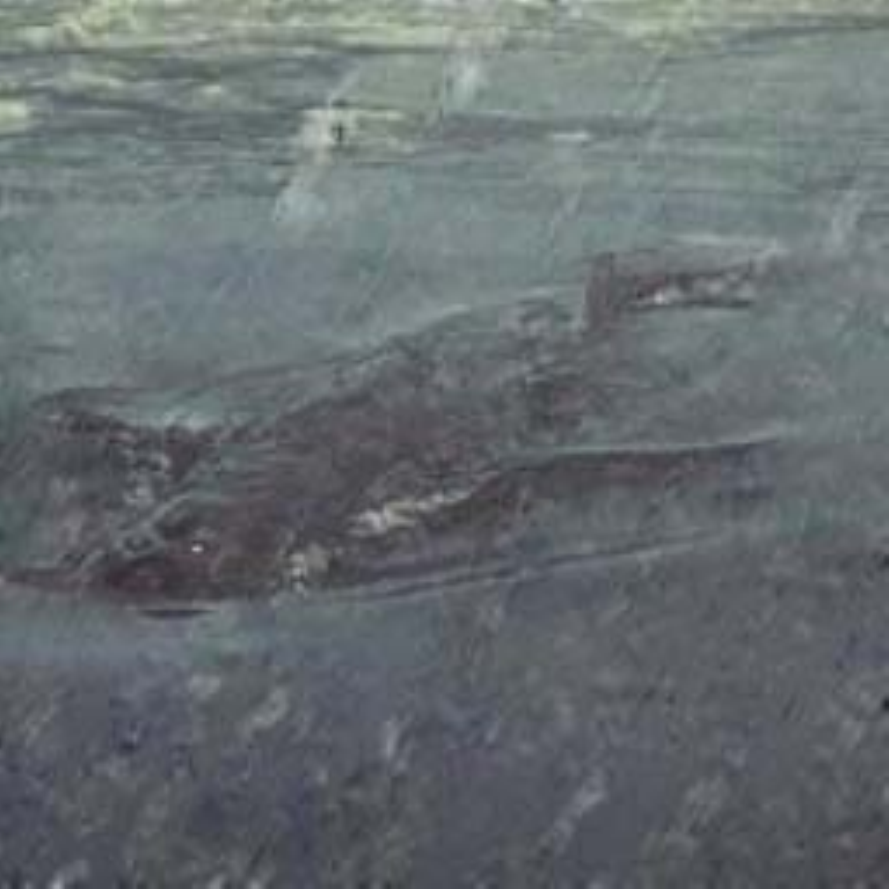}&
\includegraphics[width=\swseven]{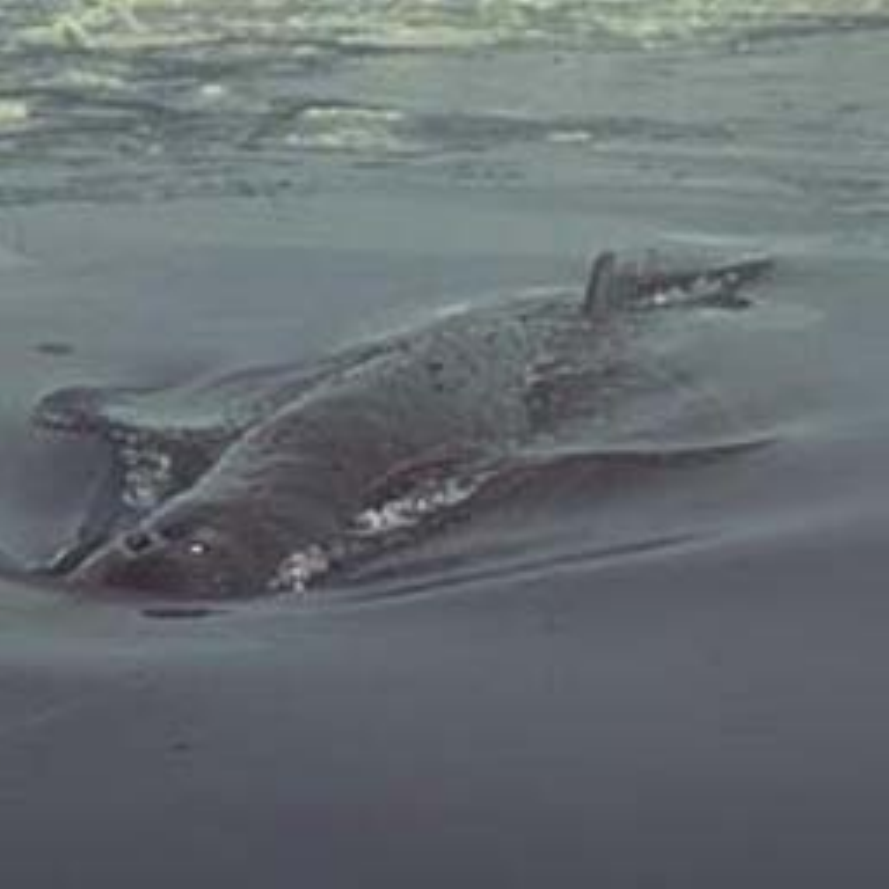}\\
(a) Input & (b) \cite{li-eccv18-recurrent} & (c) \cite{yang-pami19-joint} & (d) \cite{li-cvpr19-heavy} & (e) \cite{wang-cvpr19-spatial} & (f) Ours & (g) GT \\
\end{tabular}
\end{center}
\caption{Qualitative comparisons of selected methods and our method on synthetic rainy images. (a) Input rainy images. (b)-(f) Deraining results of RESCAN \cite{li-eccv18-recurrent}, JORDER \cite{yang-pami19-joint}, PYM+GAN \cite{li-cvpr19-heavy}, SPANet \cite{wang-cvpr19-spatial} and our method. (g) Ground truth. These two samples are two failure cases of the state-of-the-art methods.}
\label{fig:synvisual}
\end{figure*}

\renewcommand{\tabcolsep}{10pt}
\begin{table*}[t]
\centering
\caption{Averaged time cost of comparison methods with a fixed image size of $512 \times 512$.}
\begin{tabular}{cccccccc}
\hline
\hline
 Methods  & \cite{li-eccv18-recurrent}  & \cite{yang-pami19-joint} & \cite{li-cvpr19-heavy} & \cite{wang-cvpr19-spatial} & Ours \\ \hline
 Time  & $0.47s$  & $1.39s$ & $0.45s$ & $0.66s$ & $0.03s$ \\ \hline
 \hline
\end{tabular}
\label{tab:speed}
\end{table*}

\subsection{Dataset Constructions}

We follow \cite{li-cviu18-fast} to prepare training dataset where there are 20800 training pairs. The rainy image in each pair is synthesized with ground truth and rendered rainy layer by using screen blend mode. Our evaluation datasets consists of three parts. First, we randomly select 100 images from each dataset in \cite{zhang-cvpr18-density,li-eccv18-recurrent,fu-cvpr17-removing,yang-cvpr17-deep}, which brings 400 images in total and named as Rain-I. Second, we synthesize 400 images\footnote{http://www.photoshopessentials.com/photo-effects/rain/} where the synthetic rainy images possess apparent vapor, which is named as Rain-II. Third, we follow the real-world dataset ~\cite{wang-cvpr19-spatial} and name it as Rain-III. The real-world rainy images are collected from either existing works or Internet data. The independence between our training and testing datasets ensures the generalization of proposed method.

\subsection{Ablation Studies}

Our network consists of SNet, ANet, and VNet. We show how these networks work together to gradually improve image restoration results. We first remove ANet and VNet. The atmosphere light is estimated via the simplified condition illustrated in Sec.~\ref{sec:pretrainanet} to train SNet. This configuration is denoted as $C_1$. On the other side, we incorporate a pretrained ANet and use its output for SNet training, which is denoted as $C_2$. Also, we perform joint training of ANet and SNet, which is denoted as $C_3$. Finally, we jointly train ANet, SNet, and VNet where ANet and SNet are initialized with pretrained models. This configuration is the whole pipeline of our method.

Fig. \ref{fig:abla} and Table \ref{tab:abla} show the qualitative and quantitative results. We observe that the results from $C_2$ are of higher quality than those from $C_1$. The higher quality indicates that estimating atmosphere light in ideal condition is not stable for effectively image restoration as shown in Fig. \ref{fig:abla} (b). Compared to $C_2$, the results from $C_3$ are more effective to remove rain streaks, which indicates the importance of joint training. However, the vapors are not well removed in the results from $C_3$. In comparison, by adding VNet to model vapors, we observe haze is further reduced in Fig. \ref{fig:abla} (e). The numerical evaluations in Table \ref{tab:abla} also indicate the effectiveness of joint training and vapor modeling.

\subsection{Evaluations with State-of-the-art}
We compare our method with existing deraining methods on three rain datasets (i.e., Rain-I, Rain-II, and Rain-III). The comparisons are categorized as numerical and visual evaluations. The details are presented in the following:

\subsubsection{Quantitative evaluation.} Table \ref{tab:quanti} shows the comparison to existing deraining methods under PSRN and SSIM metrics. Overall, our method achieves favorable results. The PSNR of our method is about 4 dB higher than \cite{wang-cvpr19-spatial} on Rain-II dataset. In Table \ref{tab:quanti-fid}, we show the evaluations under the FID metric. This comparison shows that our method achieves lowest FID scores on all three datasets, which indicates that our results resemble most to the ground truth images. The time cost of online inference of comparison methods is shown in Table \ref{tab:speed}. Our method is able to produce results efficiently.

\subsubsection{Qualitative evaluation.} We show visual comparison from aspects of synthetic data and real-world data. Fig. \ref{fig:synvisual} shows two synthetic rain images where existing methods are able to restore effectively. In comparison, our method is effective to remove both rain streaks and vapors.

Besides synthetic evaluations, we show visual comparisons on real-world images in Fig. \ref{fig:pracvisual}. When the rain streaks are heavy as shown on the first row of (a), existing methods do not remove these streaks completely. When the rain streaks are mild as shown on the fourth row, all the comparison methods are able to remove their appearance. When the streak edges are blur as shown on the second row, the results of RESCAN and ours are able to faithfully restore while PYM+GAN tends to change the whole color perceptions in Fig. \ref{fig:pracvisual}(d). Meanwhile, there are artifacts and blocking effects appear on the third row of Fig. \ref{fig:pracvisual}(d). The limitations also arise in JORDER and SPANET where details are missing shown in Fig. \ref{fig:pracvisual}(c) and heavy rain streaks remain in Fig. \ref{fig:pracvisual}(d). Compared to existing methods, our method is able to effectively model both rain streaks and vapors. By jointing training of three subnetworks, the parameters of our visual model is accurately predicted and produces visually pleasing results.

\renewcommand{\tabcolsep}{1pt}
\def\swsix{0.16\linewidth}
\begin{figure*}[t]
\begin{center}
\begin{tabular}{cccccc}
\includegraphics[width=\swsix]{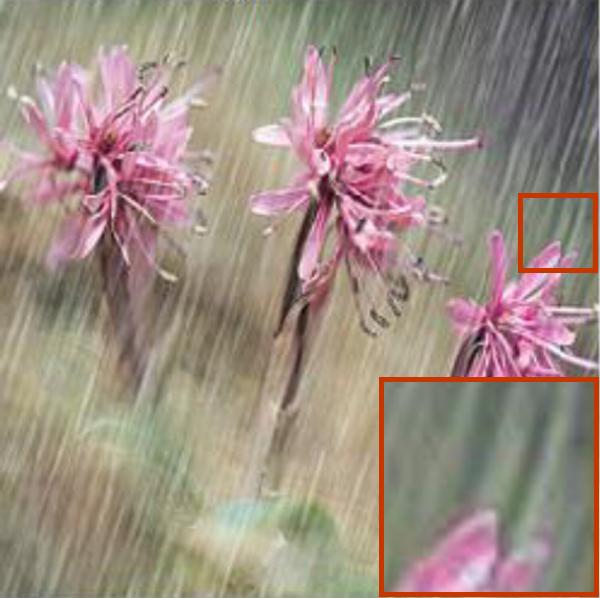}&\includegraphics[width=\swsix]{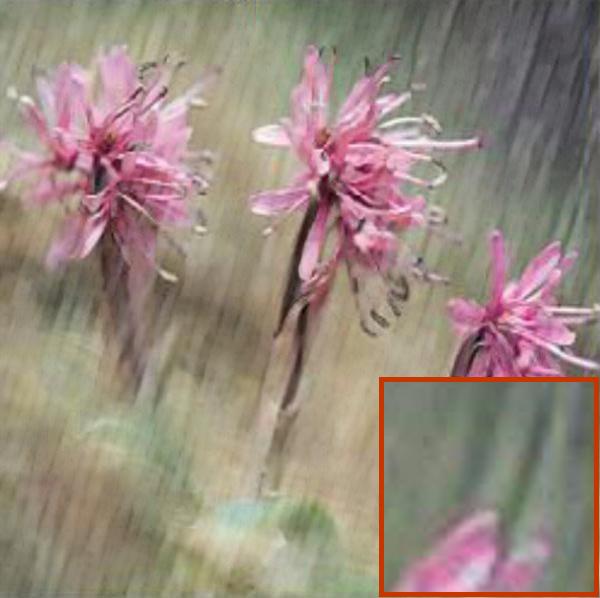}&\includegraphics[width=\swsix]{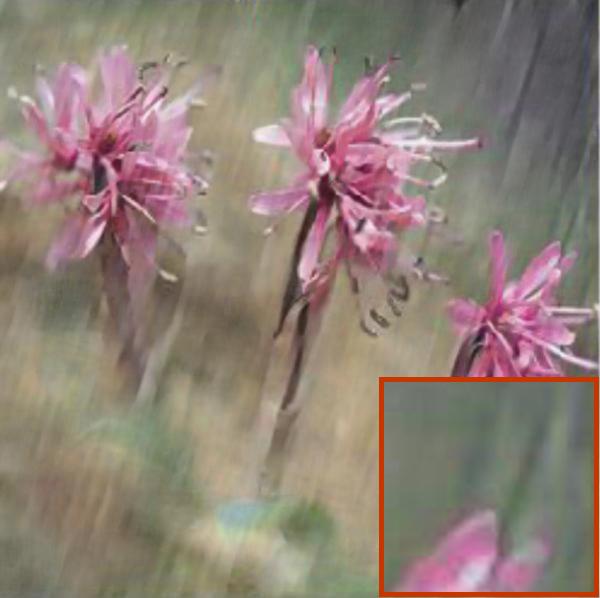}&\includegraphics[width=\swsix]{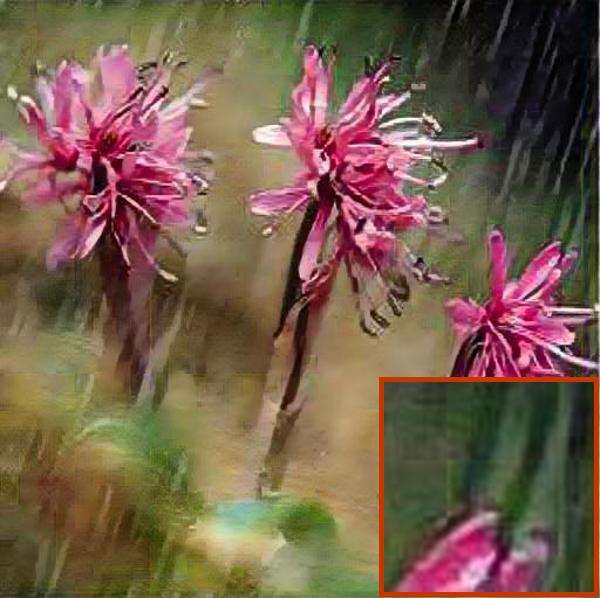}&\includegraphics[width=\swsix]{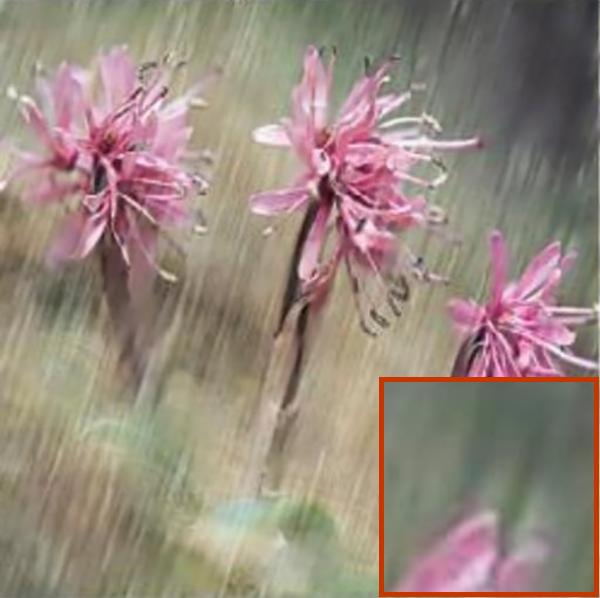}&\includegraphics[width=\swsix]{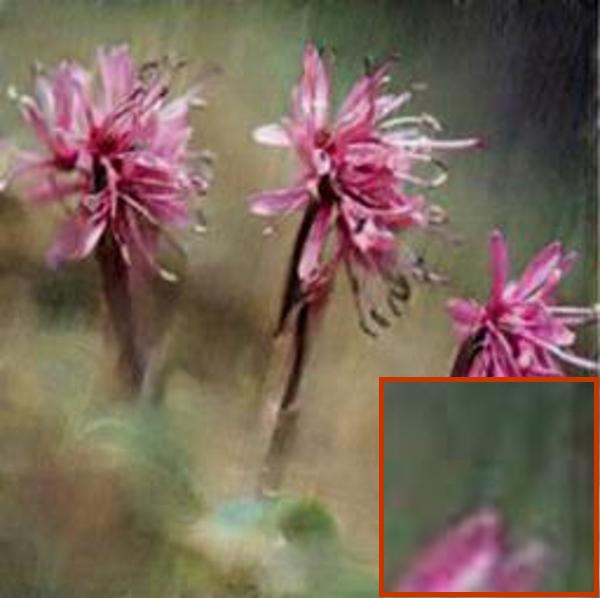}\\
\includegraphics[width=\swsix]{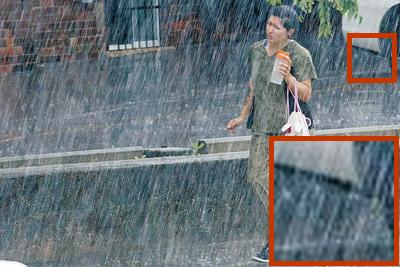}&\includegraphics[width=\swsix]{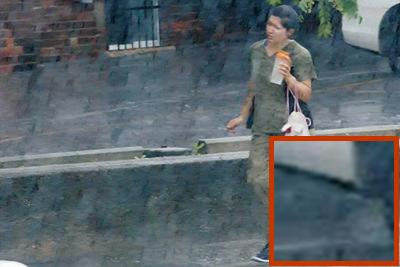}&\includegraphics[width=\swsix]{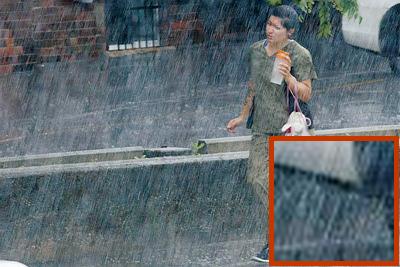}&\includegraphics[width=\swsix]{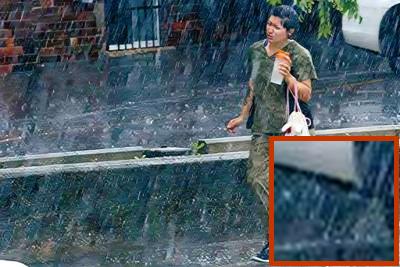}&\includegraphics[width=\swsix]{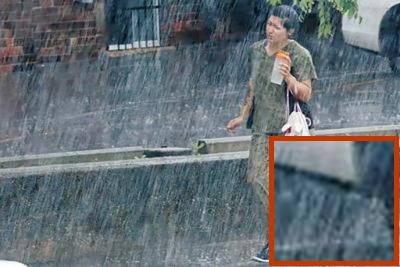}&\includegraphics[width=\swsix]{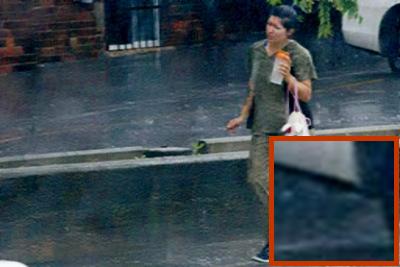}\\
\includegraphics[width=\swsix]{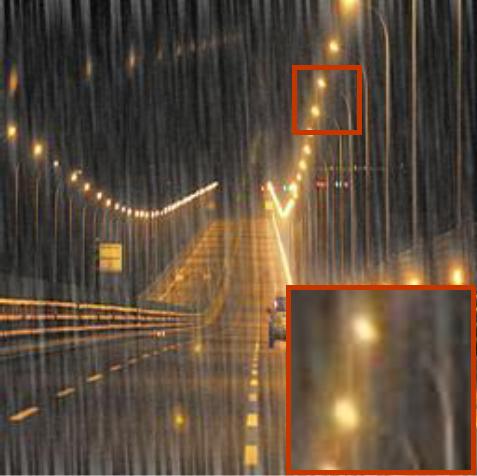}&\includegraphics[width=\swsix]{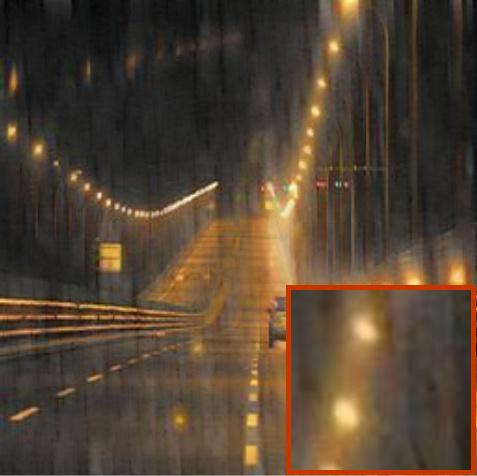}&\includegraphics[width=\swsix]{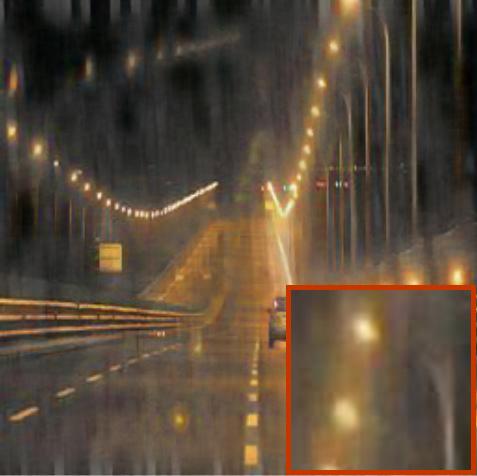}&\includegraphics[width=\swsix]{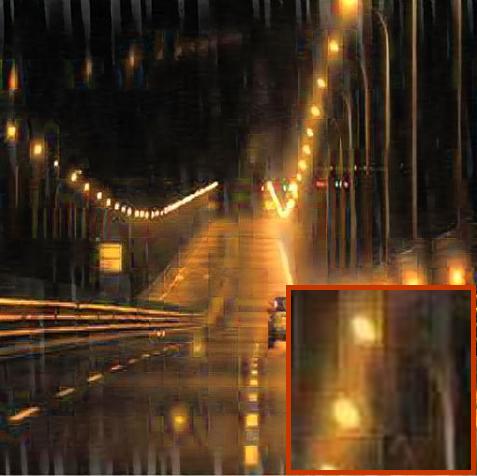}&\includegraphics[width=\swsix]{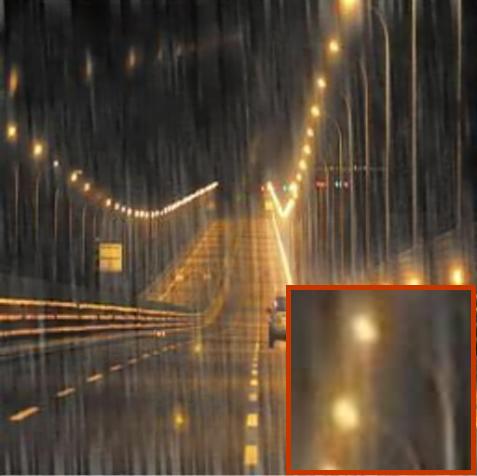}&\includegraphics[width=\swsix]{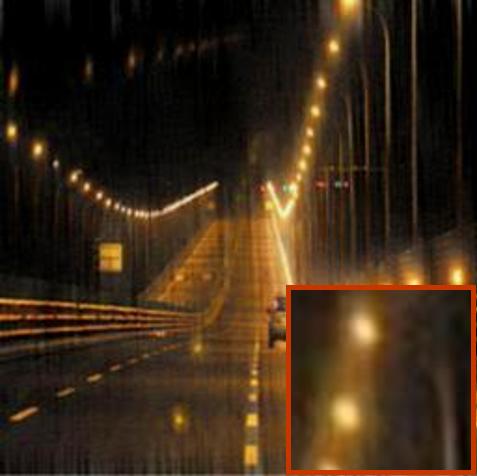}\\
\includegraphics[width=\swsix]{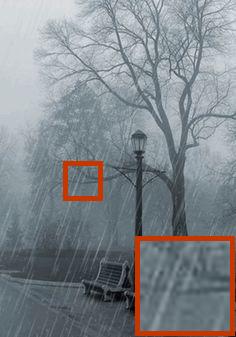}&\includegraphics[width=\swsix]{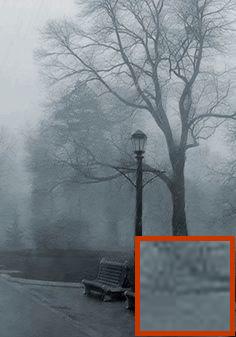}&\includegraphics[width=\swsix]{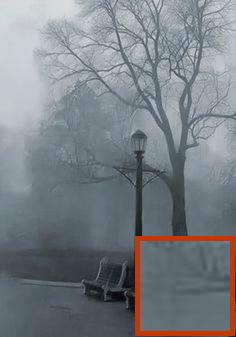}&\includegraphics[width=\swsix]{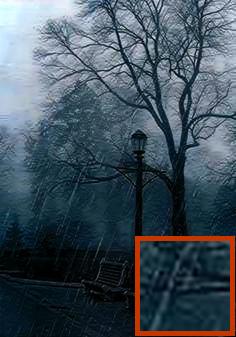}&\includegraphics[width=\swsix]{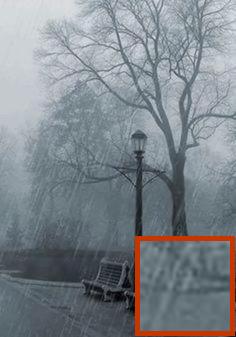}&\includegraphics[width=\swsix]{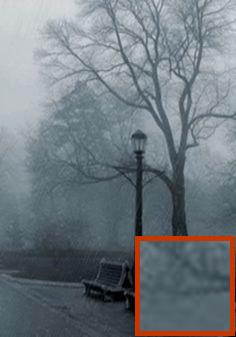}\\
(a) Input & (b) \cite{li-eccv18-recurrent} & (c) \cite{yang-pami19-joint} & (d) \cite{li-cvpr19-heavy} & (e) \cite{wang-cvpr19-spatial} & (f) Ours \\
\end{tabular}
\end{center}
\caption{Qualitative comparisons of comparison methods on real-world rainy images. Input images are shown in (a). Deraining results of RESCAN \cite{li-eccv18-recurrent}, JORDER \cite{yang-pami19-joint}, PYM+GAN \cite{li-cvpr19-heavy}, SPANet \cite{wang-cvpr19-spatial} and our method are shown from (b) to (f), respectively.}
\label{fig:pracvisual}
\end{figure*}

\section{Concluding Remarks}

\begin{flushleft}
  ``Rain is grace; rain is the sky condescending to the earth; without rain, there would be no life.''
\end{flushleft}
\begin{flushright}
  --- John Updike
\end{flushright}

Rain nourishes daily life except visual recognition systems. Recent studies on rain image restoration propose models to calculate background images according to rain image formations. A limitation occurs that the appearance of rain consists of rain streaks and vapors, which are entangled with each other in the rain images. We rethink rain image formation by formulating both rain streaks and vapors as transmission medium. We propose two networks to learn transmission maps and atmosphere light that constitute rain image formation. These essential elements in the proposed model are effectively learned via joint network training. Experiments on the benchmark dataset indicate the proposed method performs favorably against state-of-the-art approaches.

{\flushleft \bf Acknowledgement.}
This work was supported by NSFC (60906119) and Shanghai Pujiang Program.

%
%
\bibliographystyle{splncs04}
\bibliography{egbib}
\end{document}